\documentclass[fleqn,10pt]{wlscirep}
\usepackage[utf8]{inputenc}
\usepackage[T1]{fontenc}

\usepackage{hyperref}       
\usepackage{url}            
\usepackage{booktabs}       
\usepackage{amsfonts}       
\usepackage{nicefrac}       
\usepackage{microtype}      
\usepackage{lipsum}		
\usepackage{graphicx}
\usepackage{subcaption}
\usepackage{graphicx}
\usepackage{xcolor} 
\usepackage{tikz}
\usetikzlibrary{shapes.geometric, arrows, positioning, automata,positioning,fit,backgrounds}
\tikzstyle{process} = [rectangle, minimum width=2cm, minimum height=1cm, text centered, draw=black, fill=white!30]
\tikzstyle{sum} = \tikzstyle{sum} = [draw, circle, minimum size=.5cm]
\tikzstyle{arrow} = [thick,->,>=stealth]

\usepackage{amsmath}

\usepackage{bm}
\usepackage{units}
\usepackage{float}
\usepackage{dblfloatfix}
\usepackage{algorithm}
\usepackage{algpseudocode}
\makeatother
\usepackage[T1]{fontenc}
\usepackage{caption}
\usepackage{sidecap}

\usetikzlibrary{fit,calc}
\newcommand*{\tikzmk}[1]{\tikz[remember picture,overlay,] \node (#1) {};\ignorespaces}
\newcommand{\boxit}[1]{\tikz[remember picture,overlay]{\node[yshift=0pt,fill=#1,opacity=.25,fit={(A)($(B)+(.9\linewidth,.5\baselineskip)$)}] {};}\ignorespaces}
\colorlet{yellow}{yellow!100}
\colorlet{blue}{cyan!60}
\usepackage{todonotes}

\title{Lamarck's Revenge: Inheritance of Learned Traits Can Make Robot Evolution Better}

\author[1,*]{Jie Luo}
\author[1]{Karine Miras}
\author[2]{Jakub Tomczak}
\author[1]{Agoston E. Eiben}
\affil[1]{Vrije Universiteit Amsterdam, Amsterdam, The Netherlands}
\affil[2]{Eindhoven University of Technology, Eindhoven, The Netherlands}
\affil[*]{j2.luo@vu.nl}

\begin{abstract}
Evolutionary robot systems offer two principal advantages: an advanced way of developing robots through evolutionary optimization and a special research platform to conduct what-if experiments regarding questions about evolution. Our study sits at the intersection of these. We investigate the question ``What if the 18th-century biologist Lamarck was not completely wrong and individual traits learned during a lifetime could be passed on to offspring through inheritance?'' We research this issue through simulations with an evolutionary robot framework where morphologies (bodies) and controllers (brains) of robots are evolvable and robots also can improve their controllers through learning during their lifetime. Within this framework, we compare a Lamarckian system, where learned bits of the brain are inheritable, with a Darwinian system, where they are not. Analyzing simulations based on these systems, we obtain new insights about Lamarckian evolution dynamics and the interaction between evolution and learning. Specifically, we show that Lamarckism amplifies the emergence of `morphological intelligence', the ability of a given robot body to acquire a good brain by learning, and identify the source of this success: `newborn' robots have a higher fitness because their inherited brains match their bodies better than those in a Darwinian system. 

\end{abstract}
\begin{document}

\flushbottom
\maketitle
%
%

\section*{Introduction}
Evolutionary robotics (ER) is a research field that applies Evolutionary algorithms (EAs) to design and optimize the body, the brain, or both, for simulated and real autonomous robots [\cite{nolfi2000evolutionary, doncieux2015evolutionary}]. It is a promising area with a powerful rationale: as natural evolution has produced successful life forms for practically all possible environmental niches on Earth, it is plausible that artificial evolution can produce specialized robots for various environments and tasks. 

Early studies in ER explored the evolution of the controller (brain) only, while the morphologies (bodies) were fixed [\cite{Floreano1998,Capi2008}]. A holistic approach - the conjoint evolution of morphology and controller - was introduced by Karl Sims in his seminal work with virtual creatures [\cite{sims1994evolving}]. Since then, this approach has been increasingly investigated in the literature [\cite{auerbach2012relationship, Weel2014, Lipson2016, Cheney2018, Miras-Eiben-2019, Hockings2020, Stensby2021, Medvet2021, Nygaard2021}]. While the inclusion of body evolution was a fundamental step toward complex robotic intelligence, there is (at least) one more layer that should be taken into consideration: learning[\cite{Nolfi1999}]. Learning allows to fine-tune the coupling between body and brain and also allows for more degrees of freedom that account for environmental changes.  The majority of such research consists of applying learning algorithms to the evolvable brains of robots with fixed bodies [\cite{Floreano1996, Harvey1997, Schembri2007, Sproewitz2008, Bellas2010, Khazanov2014, Ruud2017, Schaff2019, Hwangbo2019, LeGoff2020}], but some have also evolved the morphologies [\cite{Lehman2011,Jelisavcic2017,Kriegman2018,Aguilar2019,Howard2019,Miras2020,Gupta2021,Goff2021a,Hart2022,Luo2022}].

Unlike classical engineering approaches, based on mathematics and physics, ER is inspired by a less understood mechanism: biological evolution. In biology, experimental research is often slowed down by the fact that evolution requires many generations with large populations of individuals whose lives may last many decades. For this reason, most of the research focuses on organisms whose life cycle is short enough to allow laboratory experiments [\cite{wiser2013long}]. ER offers a synthetic alternative approach where robots are the evolving entities that enable the experimental testing of hypotheses [\cite{long2012darwin}]. As John Maynard Smith, one of the fathers of modern theoretical biology, argued: "So far, we have been able to study only one evolving system and we cannot wait for interstellar flight to provide us with a second. If we want to discover generalizations about evolving systems, we will have to look at artificial ones."[\cite{maynard1992byte}]. ER has been already used to study some of the key issues in evolutionary biology, such as the evolution of cooperation, whether altruistic [\cite{montanier2011surviving, waibel2011quantitative}] or not [\cite{solomon2012comparison}], the evolution of communication [\cite{floreano2007evolutionary, mitri2009evolution, wischmann2012historical}], morphological complexity [\cite{bongard2011morphological, auerbach2014environmental}], and collective swarming [\cite{olson2013predator}]. 

One of the most enduring controversial matters in evolutionary biology is Lamarckism, which asserts that adaptations acquired or learned by an individual during its lifetime can be passed onto its offspring [\cite{Conway1946}]. Although this theory has been disproven by modern genetics, it is important to note that the concept of Lamarckian evolution is still a subject of debate in the scientific community, and there is no consensus on whether or not it occurs to some extent in nature [\cite{Burkhardt2013}]. For instance, one could argue that epigenetic changes [\cite{bossdorf2008epigenetics}] allow for Lamarckism to occur. Epigenetics studies patterns of gene expression that can be inherited and might remain active for multiple generations. This means that genetic expression regulated throughout the life of an organism can be transmitted to its offspring through temporary modifications to molecular structures of the DNA, but which do not change the DNA sequence itself. The link between Lamarckism and epigenetics comes from the following reasoning: the behavior and environmental exposure of an organism might induce certain epigenetic changes; these changes can be reflected directly in the phenotypic expression of the offspring; therefore, what the parents experience during their lives might directly affect the phenotype of their offspring.

By simulating Lamarckian evolution in an artificial, non-biological substrate, we can study it in a what-if fashion: What if Lamarck was right and individual traits acquired during a lifetime could be passed on to offspring through inheritance? Empirical data delivered by computer simulations can help study and analyze the evolutionary dynamics and explore the potential benefits and drawbacks of Lamarckian evolution for developing robots. Thus, on the one hand, from the robotics perspective, this can contribute to more advanced evolutionary algorithms that in turn can deliver better robotic systems, perhaps even in less time. On the other hand, from the biological perspective, this can contribute to insights into natural evolution; not Life as we know it, but Life as it could be. 

Previous research on artificial evolution combined with learning is mostly limited to Darwinian systems [\cite{Luo2022,Gupta2021,Goff2021,Nygaard2021}] and the Baldwin effect [\cite{Kriegman2018,Aguilar2019}]. The few existing studies on Lamarckian evolution can be divided into three categories. First, disembodied evolution applying an evolutionary algorithm to machine learning techniques [\cite{mingo2013,whitley1994, Holzinger2014, mikami1996, ku1999, houck1997, castillo2006, ZHANG2013, elsken2018}]. In this category, studies found that the Lamarckian mechanism quickly yields good solutions, accelerating convergence and adapting to dynamic environments but may risk converging to a local optimum and may yield different results in different domains. Second, embodied evolution of controllers for robots with fixed bodies [\cite{Nishiwaki2015,Grefenstette1991,Ruud2017}]. In this category, studies found that Lamarckian evolution is effective in improving the performance of robot controller evolution, and that the learning process reduces the negative impact of the simulation-reality gap. Finally, there is the full-blown case, embodied evolution of morphologies and controllers together in a Lamarckian manner. This is the most complex category that has hardly been studied so far with only two papers (of ourselves) we are aware of [\cite{jelisavcic2019lamarckian,jelisavcic2018morphological}]. These considered the simplest possible robot task (undirected locomotion, a.k.a. gait learning) and observed the increased efficiency and efficacy of Lamrackian evolution compared to the Darwinian counterpart. 

Importantly, all previous studies focused on establishing the advantages of Lamarckism without a deeper investigation into why and how Lamarckism delivers such benefits and to date there is hardly any knowledge about the most complex case of morphologically evolvable robots. This latter may be rooted in the difficulty in designing and implementing such a system. Technically speaking, it requires a reversible mapping between (certain segments of) the genotype and the phenotype. In particular, some features of the robot controller must be evolvable (i.e. inheritable) as well as learnable, and the traits acquired by the learning algorithm during the lifetime of a robot must be coded back to the genotype to make them inheritable. 

In this work, we investigate the effects of Lamarckism on morphologically evolvable robots. Specifically, we apply a Lamarckian system that acts upon the learning layer --what the parents learn can be inherited by the offspring-- while solving the reversible genotype-phenotype mapping problem. 
We compare the Lamarckian system to a Darwinian system [\cite{Luo2022}] in which learning occurs, but learned traits are not inherited. Both systems include body evolution, brain evolution, and learning. All properties and parameters in both systems are the same, except that the inheritance of learned traits is present only in the Lamarckian system. Specifically, we test three hypotheses:
\begin{itemize}
    \item  The Lamarckian system is more effective and efficient than the Darwinian system.
    \item  The Lamarckian system converges into superior bodies faster.
    \item  The Lamarckian system produces better `newborns' with relatively high performance even before learning takes place.
\end{itemize}

The main contributions of the present work are twofold: a) a general framework for a Lamarckian robot evolution system with a reversible genotype-phenotype mapping, and b) novel insights into the deeper effects of Lamarckism underlying the increased effectiveness and efficiency occur.


\newpage
\section*{Results}
We arrange the results around different robot features: task performance, morphology and behavior. 

\subsection*{Task Performance}
\begin{figure}
    \begin{minipage}{0.49\textwidth}
    \includegraphics[width=0.9\linewidth]{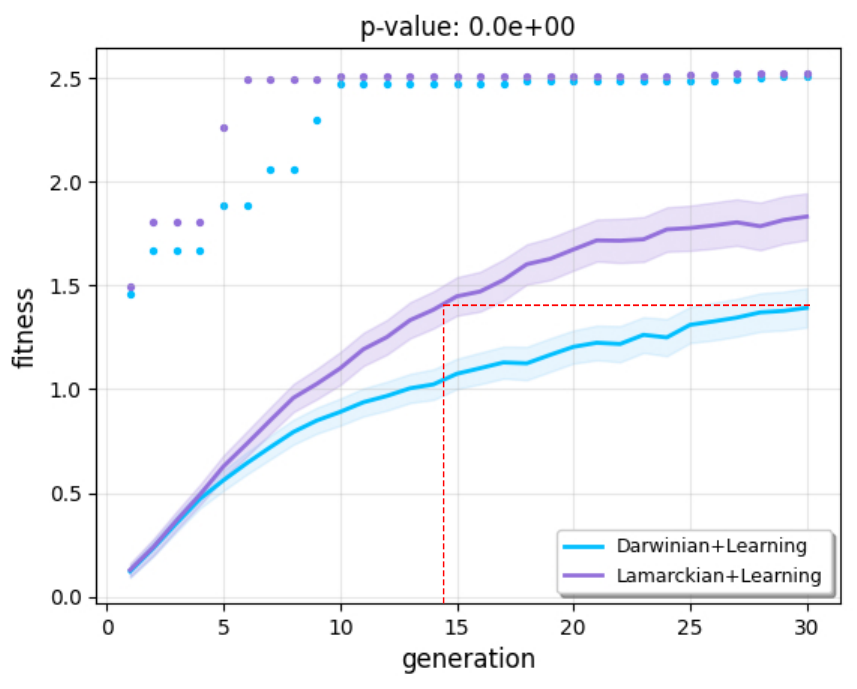}
    \caption{\label{fig:fitness_mean_avg}Mean (lines) and maximum (dots) fitness over 30 generations. The bands indicate the 95\% confidence intervals (Sample Mean $\pm$ t-value $\times$ Standard Error).}
    \end{minipage}
    \begin{minipage}{0.49\textwidth}
        \includegraphics[width=0.87\linewidth]{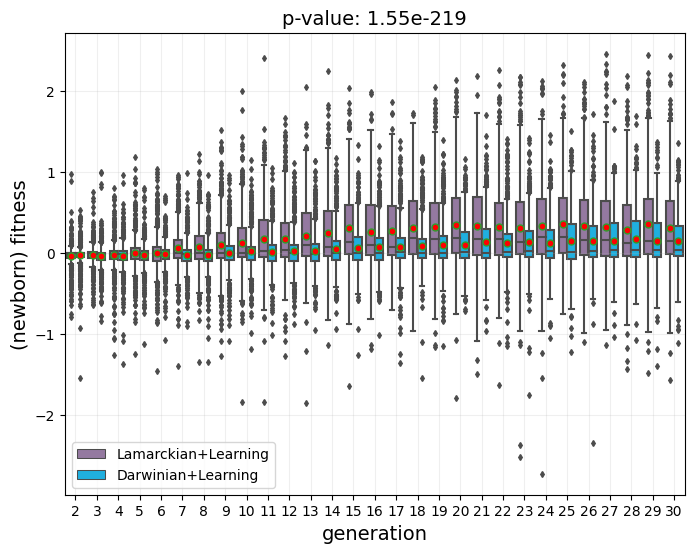}
        \caption{Distribution of newborn fitness before learning (all runs merged). Red dots indicate the mean values. The p-value in the title is the significance of the comparison between Lamarckian and Darwinian systems, including all newborns ever born (all generations and all runs merged).}
        \label{fig:newborn_fitness_before}
    \end{minipage}
\end{figure}

Robots are evolved for a point navigation task, requiring that the robot visits a sequence of target points (See the Methods section for details). Their task ability is used as the fitness function for evolution and as the reward function for the lifetime learning method, cf. Algorithm 1. 
Figure \ref{fig:fitness_mean_avg} exhibits the development of fitness over consecutive generations of the Lamarckian and the Darwinian systems. These curves show that the best robots that the Darwinian system produces reach a fitness of 2.5, but the populations produced by the Lamarckian system are significantly better - approximately 25\% higher at the end.

\begin{figure}

   \begin{subfigure}{0.49\textwidth}
    \includegraphics[width=0.9\linewidth]{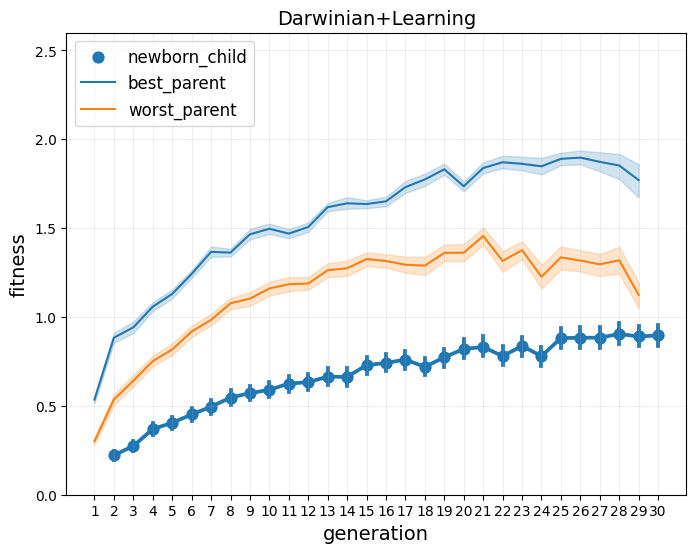}
    \caption{}
    \end{subfigure}\hfill
    \begin{subfigure}{0.49\textwidth}
        \includegraphics[width=0.9\linewidth]{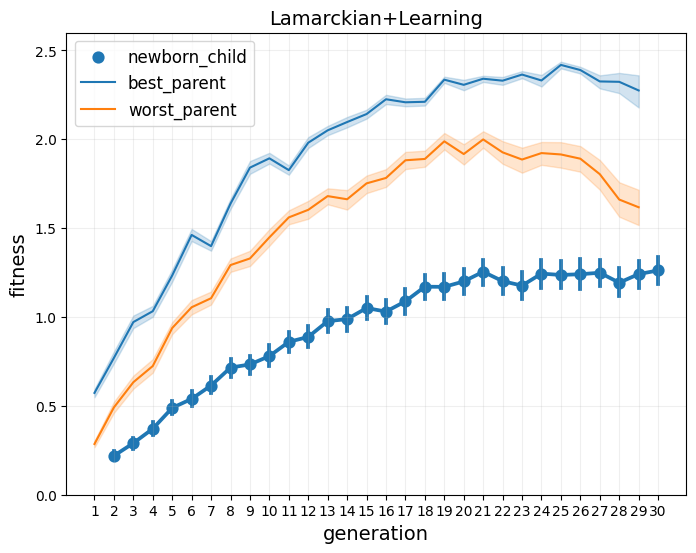}
        \caption{}
        \label{fig:lam_fitness_parents_newborns}
    \end{subfigure}
    \caption{Average fitness: comparison between newborns (after learning) and their parents (after learning) across generations.}
  \label{fig:parent_child}
\end{figure}

\begin{figure*}
     \centering
     \includegraphics[width=\linewidth]{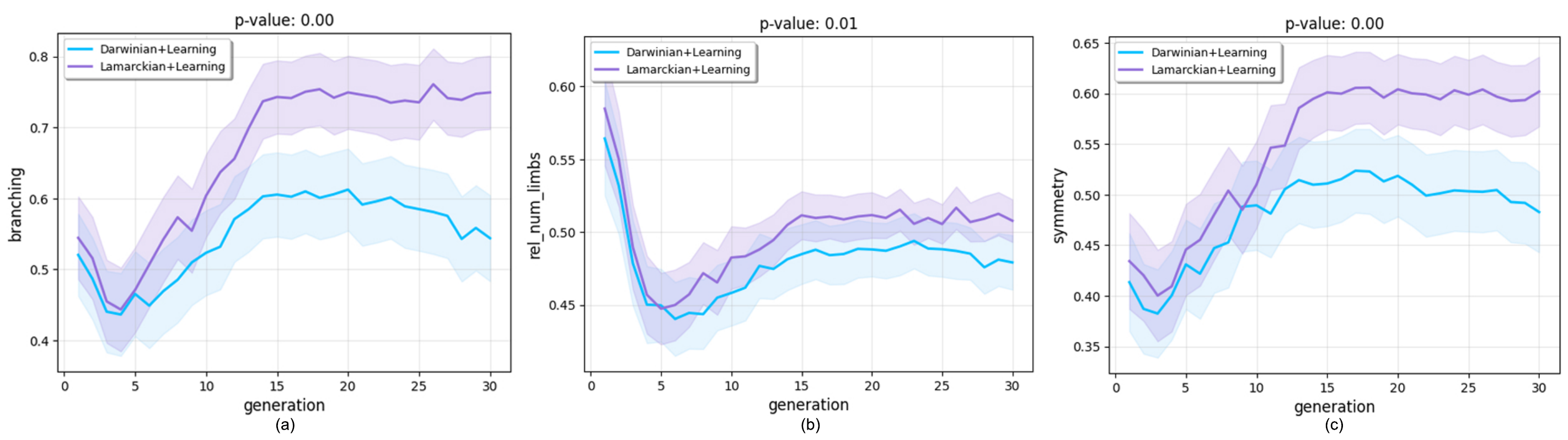}
    \caption{\label{fig:traits} Morphological traits over generations. We present the progression of their means averaged over 20 runs for the entire population. Shaded regions denote a 95\% confidence interval. The significance level after Bonferroni correction for 6 comparisons is p < 0.006.}
    \vspace{-1em}
\end{figure*}

\begin{figure*}
   \begin{minipage}{0.65\textwidth}
     \centering
     \includegraphics[width=1\linewidth]{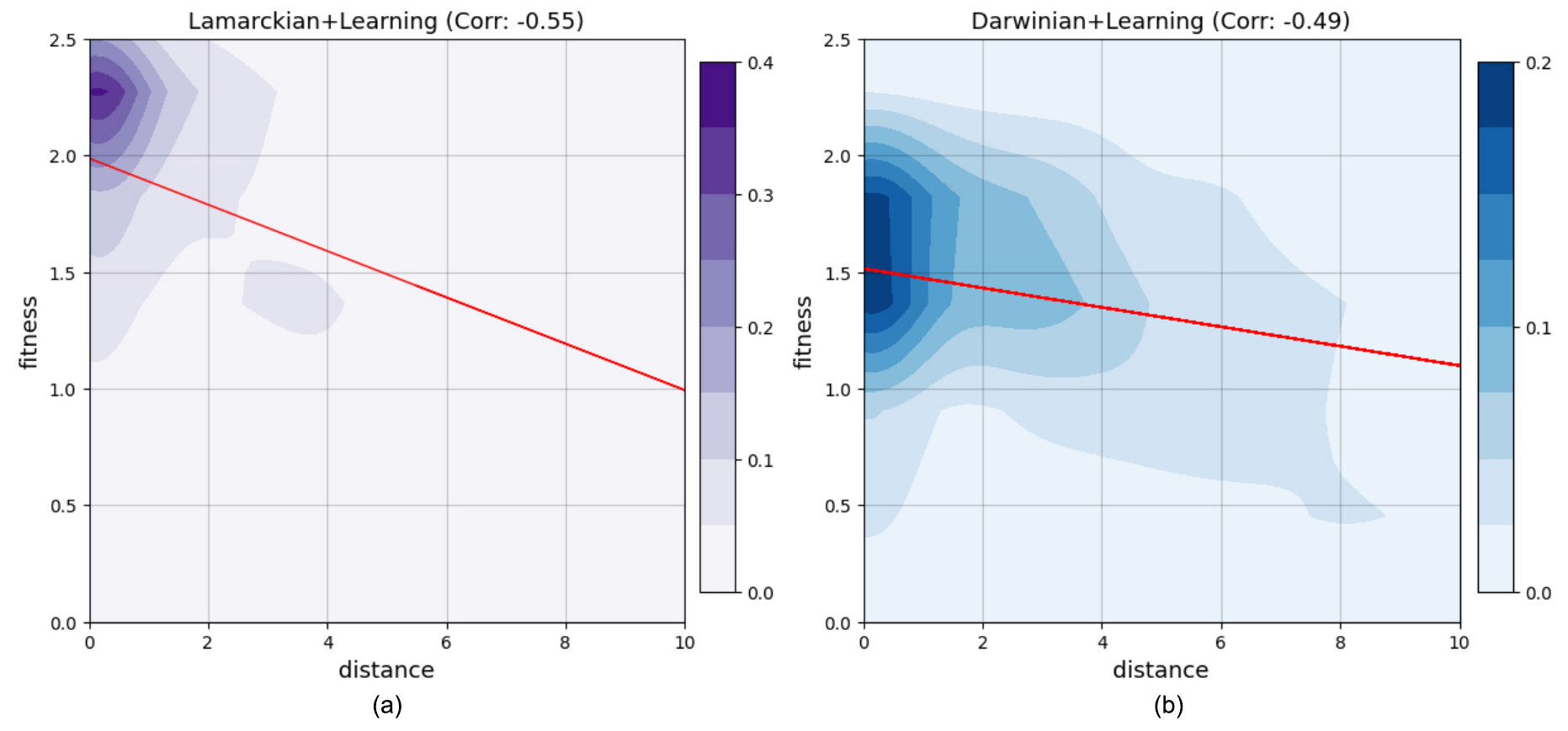}
     \caption{\label{fig:distance1}Tree-edit distance density plots. (a) is the fitness over distance for the Lamarckian system. (b) is for the Darwinian system. The darker the color, the higher density of the robots in that region. The red lines are the regression lines. The correlation efficiency rate is shown in the title of each plot.} 
   \end{minipage}\hfill
   \begin{minipage}{0.32\textwidth}
   \vspace{-14pt}
     \centering
     \includegraphics[width=1\linewidth]{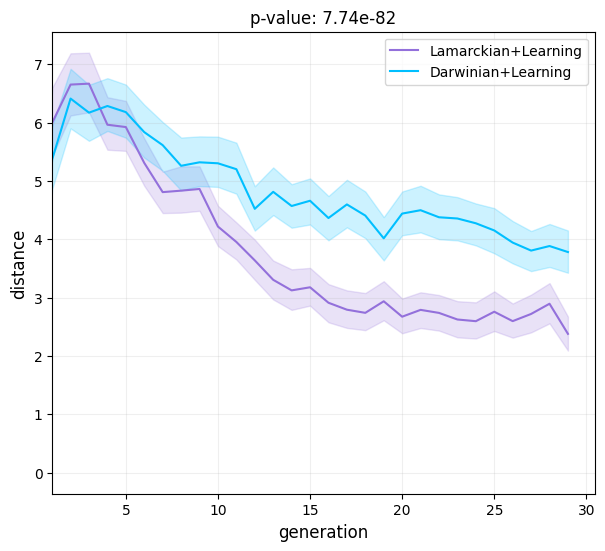}
    \caption{\label{fig:distance2} Tree-edit distance over generation. Shaded regions denote a 95\% confidence interval.}
    \end{minipage}
\end{figure*}



Figure \ref{fig:fitness_mean_avg} also demonstrates the differences in efficiency. The Lamarckian system is more efficient than the Darwinian one, as it finds the best solutions (robots with the highest fitness) much faster. Furthermore, the dotted red lines show that halfway through the run (around generation 14) the Lamarckian system has reached the quality produced by the Darwinian system only by the end of the evolutionary process. This can be seen as significant `savings' of 2,240,000 evaluations (25 offspring $\cdot$ 16 generations $\cdot$ 280 learning trials $\cdot$ 20 runs).

To investigate more closely what allows the Lamarckian system to be more effective and efficient than the Darwinian, we inspected the fitness of the robots both before and after learning. Figure~\ref{fig:parent_child} compares the fitness of the newborns after learning with the fitness of their parents also after they learned: not only are the Lamarckian parents better than the Darwinian parents, but also are the Lamarckian newborns better than the Darwinian newborns. Additionally, Figure~\ref{fig:newborn_fitness_before} shows the fitness distributions of the newborns before learning: the fitness distributions of newborns reach higher values through the Lamarckian system than through the Darwinian system. This holds for almost all generations except a few at the beginning of the search. These observations mean that not only are Lamarckian robots better after they learn, but also better immediately after they are born.

\subsection*{Robot Morphologies}

We analyze the morphological properties of the robots addressing four different aspects: the morphological traits (details about the measures can be found in [\cite{miras2018search}]), the morphological similarity between offspring and parents, the morphological diversity at each generation and the morphological intelligence. 

\subsubsection*{Morphological traits}
To investigate the morphologies generated by the Lamarckian and Darwinian evolution systems consider eight morphological traits to quantitatively analyze the evolved morphologies of all robots. 
Among these eight traits, only three of them presented significant differences (Figure \ref{fig:traits}), namely branching, number of limbs, and symmetry. Robots evolved by the Lamarckian system tend to be more symmetric and have more branches and limbs than robots evolved with the Darwinian system. Nevertheless, despite these observed differences, visual inspection of the top bodies does hardly allow for an intuitive differentiation between their shapes (Figure ~\ref{fig:best5}). Moreover, a PCA analysis using these same eight traits does not show any difference between the morphologies produced by each method. Therefore, although there is evidence for some differences in morphological traits, these differences are marginal (Figure \ref{fig:pca}).
    


\subsubsection*{Morphological similarity}

In our research, the morphological similarity is calculated as the tree-edit distance of morphological structures between each child and the fittest parent. We use the APTED algorithm, the state-of-the-art solution for computing the tree-edit distance [\cite{Pawlik2015}].


Figure \ref{fig:distance1} shows the correlation between fitness and distance. For both methods, the correlation between fitness and distance is negative. This means that the most similar the offspring is to the parent, the higher the fitness of the offspring. Importantly, this correlation is even stronger in the case of the Lamarckian system.

Furthermore, \ref{fig:distance2} shows how the average distance progresses over the generations. 
With both systems, we see pressure for reducing the distance between the offspring and parent, but this pressure is higher with the Lamarckian system. This effect is logical because it is expected that the brain of a parent would be a better match to a body similar to its own body.


 

\subsubsection*{Morphological diversity}
Morphological diversity is the morphological variety of each population using tree-edit distance. It is calculated as the average distance of the difference between any two robots d(x,y) at each generation.

Figure \ref{fig:diversity} illustrates a notable trend: the morphological diversity of the Lamarckian system declines at a more rapid rate compared to the Darwinian system. This observation strongly suggests that the Lamarckian system converges into superior bodies at a faster pace.

\subsubsection*{Morphological intelligence}

Morphology influences how the brain learns. Some bodies are more suitable for the brains to learn than others. How well the brain learns can be empowered by a better body. Therefore, we define the intelligence of a body as a measure of how well it facilitates the brain to learn and achieve tasks. 
 
We quantify morphological intelligence by the delta of the learning delta of each method, being the learning delta of the evolved body minus the learning delta of the fixed body, whereas the learning delta, being the fitness value after the parameters were learned minus the fitness value before the parameters were learned. 
 
 To verify the presence of morphological intelligence, we conducted an additional experiment. First, we evolved robot brains for fixed bodies, given that these bodies were the same initial (random) bodies produced by the main experiments. Additionally, learning was carried out just like in the main experiments. Second, we calculated the learning delta of each individual as the fitness after learning minus the fitness before learning. Finally, we compared the average learning delta produced by the experiments with fixed bodies (described above) with the learning delta from the main experiments (evolvable bodies).
 
\begin{figure*}
   \begin{minipage}{0.48\textwidth}
     \centering
         \includegraphics[width=0.9\textwidth]{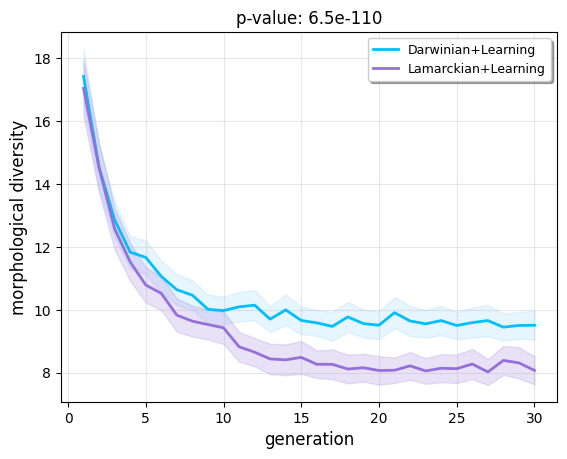}
        \caption{\label{fig:diversity} Morphological diversity of two systems averaged over 20 runs at each generation. The P-value shows a significant difference between the means diversity value of the two systems. The bands indicate the 95\% confidence intervals. }
   \end{minipage}\hfill
   \begin{minipage}{0.48\textwidth}
     \centering
     \includegraphics[width=0.9\linewidth]{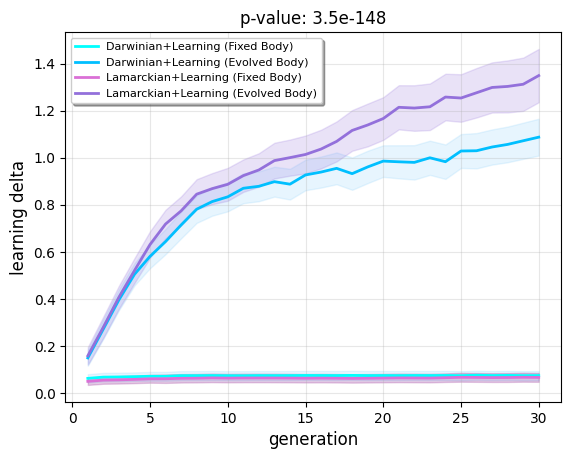}
    \caption{\label{fig:learning delta}Progression of the learning delta throughout evolution averaged over 20 runs. The P-value shows a significant difference between the mean learning delta of two systems (evolved body). The bands indicate the 95\% confidence intervals.}
    \end{minipage}
\end{figure*}

 In Figure \ref{fig:learning delta}, we see that the average learning deltas of both methods with evolved bodies grow steadily across the generations which indicates that lifetime learning leads the evolutionary search towards morphologies with increasing learning potential. The average learning delta of fixed bodies, on the other hand, grows so little that it can hardly be seen when included on the same axis of the evolvable bodies experiments. For the Lamarckian system and Darwinian system, the learning delta with evolvable bodies is around 1885\% and 1305\% higher than when using fixed bodies respectively. Moreover, this delta is around 30\% greater for the Lamarckian system than the Darwinian system. 
 
 The plot also illustrates that the delta of learning delta between evolved body and fixed body for each method is growing across generations which demonstrates that this learning delta growth results (specifically) from morphological intelligence, and not simply from the presence of evolution.  

\begin{SCfigure}[][h!]
  \centering
  \includegraphics[width=0.67\linewidth]{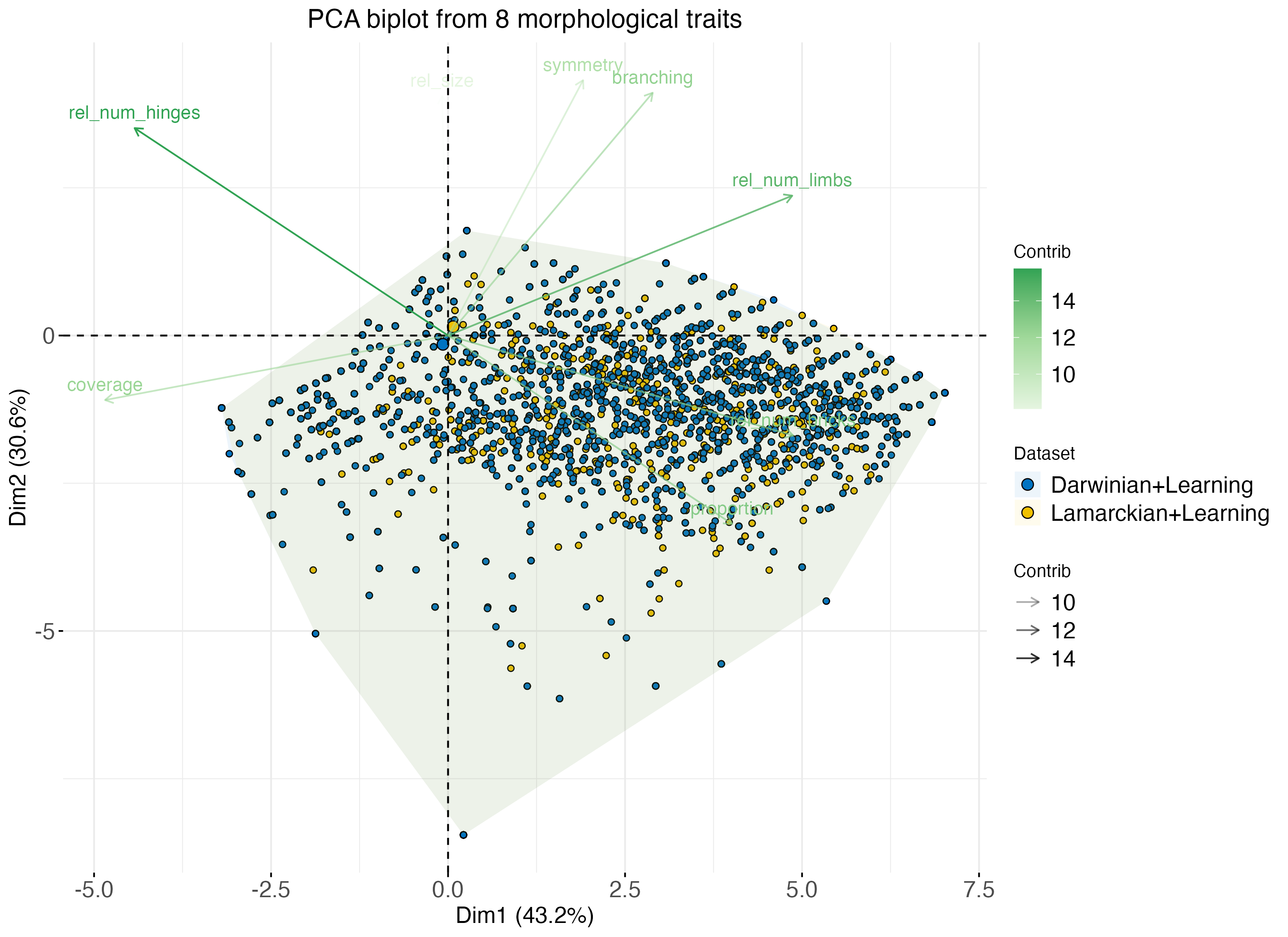}
    \captionsetup{justification=justified}
  \caption{Principal Component Analysis (PCA) biplot showing the distribution of samples based on 8 morphological traits in datasets of both methods. Each point represents a robot sample, and the plot displays the first two principal components (Dim1 and Dim2), which explain 43.2\% and 30.6\% of the total variance, respectively. Furthermore, the biplot displays the variables (morphological traits) as arrows, representing their contribution to the principal components. Traits pointing in similar directions are co-regulated or have similar expression patterns across the samples.}
  \label{fig:pca}
\end{SCfigure}

\begin{figure*}[h!] 
  \centering
     \begin{subfigure}[b]{0.49\textwidth}
         \centering
         \includegraphics[width=0.99\textwidth]{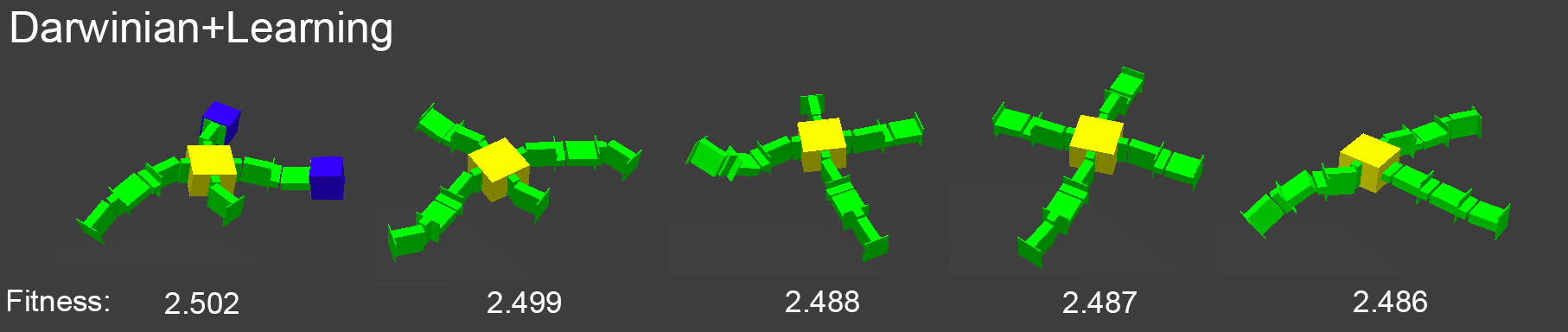}
         \caption{}
     \end{subfigure}
     \hfill
     \begin{subfigure}[b]{0.49\textwidth}
         \centering
         \includegraphics[width=0.99\textwidth]{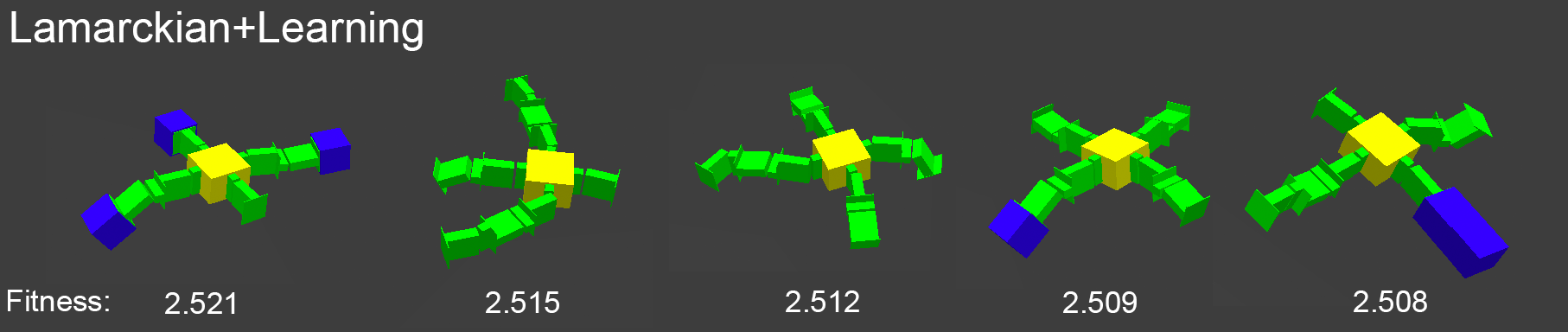}
         \caption{}
     \end{subfigure}
  \caption{The 5 best robots produced by both methods with their fitnesses.}
  \label{fig:best5}
\end{figure*}

\subsection*{Robot Behavior}
\begin{figure*}
    \centering
    \begin{subfigure}[b]{0.48\textwidth}
        \centering
        \includegraphics[width=0.9\textwidth]{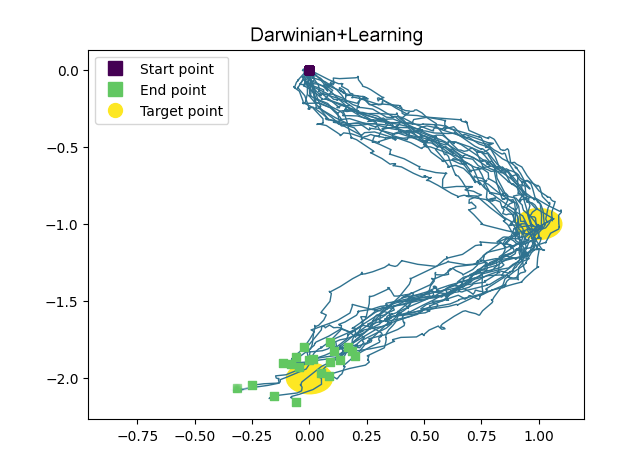}
    \end{subfigure}
    \hfill
    \begin{subfigure}[b]{0.48\textwidth}  
        \centering 
        \includegraphics[width=0.9\textwidth]{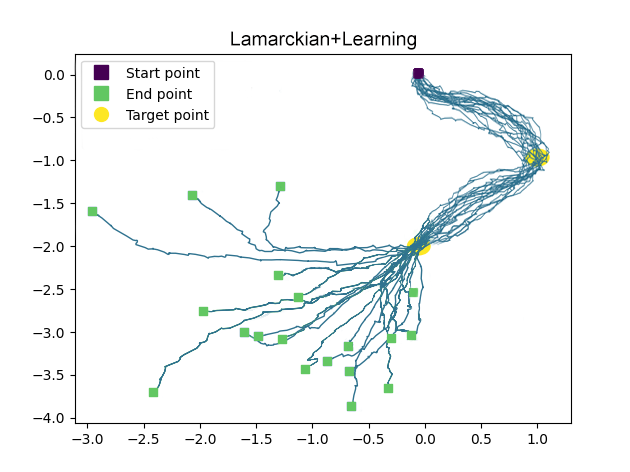}
    \end{subfigure}
    \caption{Trajectories of the best 20 robots from both methods in the point navigation task. The purple square is the starting point. Two yellow circles are the target points which robots aim to go through. The blue lines are the trajectories of robots ending at the green squares.}
    \label{fig:trajectories}
\end{figure*}

To obtain a better understanding of the robots' behaviour, we visualize the trajectories of the 20 best-performing robots from both methods in the last generation across all runs. Figure \ref{fig:trajectories} shows that all robots from the Lamarckian system reached the two target points much earlier than the ones from the Darwinian system. This can be concluded because after reaching the target, they still have time to keep moving further from the target.

\section*{Discussion}
This investigation exceeds existing studies about Lamarckism in (simulated) robot evolution systems. It is not limited to evolving brains for fixed bodies, but considers Lamarckism in the most interesting case that has hardly been studied before, where morphologies and controllers both undergo evolution. A key feature of the system is the invertible genotype-phenotype mapping regarding the robot brains. This is an important prerequisite for making learned traits inheritable, because learning always acts on the phenotypes, after `birth'. If the genotype-phenotype mapping is invertible, then the newly learned traits that were not present in the robot at `birth' can be coded back to its genotype before it reproduces. In turn, this makes the learned traits inheritable, thus evolvable. Our solution is developed for modular robots whose body configuration is evolvable. While in our current system, the number of different modules is limited, the principle behind our design is generic, and applicable for robots with more different modules, if only the controller architecture can be derived from the morphology, effectively parameterizing the search space of possible robot brains.    

The first set of our findings reconfirms earlier results about the increased efficiency and efficacy of Lamarckian evolution. Specifically, we showed that the Lamarckian system reaches the top fitness levels of the Darwinian system with just half of the effort (Figure \ref{fig:fitness_mean_avg}). Additionally, it presents a higher overall efficacy: the average fitness of the final populations is 25\% higher when using the Lamarckian system than the Darwinian system. Although previous work, from ourselves, has published similar results, the task used in \cite{jelisavcic2019lamarckian,jelisavcic2018morphological} was very simple (undirected gait learning). Here we move the front of applicability, showing that Lamarckian evolution is also superior in practically more relevant cases.

We also showed that, although the best morphologies found by the Lamarckian and Darwinian systems are similar to each other (Figure \ref{fig:pca}, \ref{fig:best5}), the two systems differ in the morphologies created during evolution. In particular, the Lamarckian system produced offspring more similar to the parents (Figure \ref{fig:distance1}, \ref{fig:distance2}) and converged into superior bodies faster (Figure \ref{fig:diversity}). As a consequence, the learning applied to the brains of the given bodies sped up in finding the optimal brains (Figure \ref{fig:learning delta}). 

One new insight about Lamarckian evolution was revealed by using the notion of the learning delta, the increase of performance achieved by learning after birth. This is a specific concept for morphological robot evolution combined with learning. In such systems, a `newborn' robot has a body and a brain, hence its fitness can be measured immediately. Additionally, fitness can be measured also after the learning process and the difference can be calculated. Figure \ref{fig:learning delta} indicates the emergence of morphological intelligence: over the course of evolution, the bodies are becoming better learners in the Darwinian as well as the Lamarckian system. However, this effect was intensified with Lamarckism at 30\%, therefore, Lamarckism is deemed superior in the joint evolution of bodies and brains. 


Finally, we demonstrated that the newborns produced by the Lamarckian system are better even before the learning process takes place. Despite sounding logical, this observation is not obvious. One reasonable premise to expect Lamarckism to be beneficial is: if parents provide an initial load of `cognitive resources' to their offspring, this should facilitate the offspring's initial behavior in contrast to starting from scratch. However, such an assumption could turn out not to be true. For instance, it could be that the newborns are not particularly better and that the main role of Lamarckism is acting as a smart initialization operator, which later on leads to better learning. Therefore, the main contribution of the current work is demonstrating where these benefits are coming from. This work contrasts with all previous literature, which explored the conditions in which Lamarckism could be beneficial but did not provide insights about where these benefits come from.


Importantly, one limitation of the present study is the use of small population sizes due to the high computational costs involved. Moreover, we experimented only with simulated robots; applying the Lamarckian system to physical robots and testing their performance in diverse environments would provide valuable insights into the practical applications of our findings. Finally, while Lamarckism has presented benefits in a scenario where the environment is static, it would be interesting to see whether it would still be beneficial when environmental conditions change rapidly or frequently. In such a scenario, Lamarckism could perhaps be counter-productive, leading to over-fitting to a reality that is no longer true when the offspring is born or the opposite. With these challenges in mind, we consider that future work should address Lamarckism in the context of changing environments.

\section*{Methods}
\subsection*{Robot Morphology (Body)}
\subsubsection*{Body Phenotype}
The phenotype of the body is a subset of RoboGen's 3D-printable components [\cite{Auerbach2014}]: a morphology consists of one core component, one or more brick components, and one or more active hinges. The phenotype follows a tree structure, with the core module being the root node from which further components branch out. Child modules can be rotated 90 degrees when connected to their parent, making 3D morphologies possible. The resulting bodies are suitable for both simulation and physical robots through 3D printing.


\subsubsection*{Body Genotype}
The phenotype of bodies is encoded in a Compositional Pattern Producing Network (CPPN) which was introduced by Stanley [\cite{Stanley2007}] and has been successfully applied to the evolution of both 2D and 3D robot morphologies in prior studies as it can create complex and regular patterns. The structure of the CPPN has four inputs and five outputs. The first three inputs are the x, y, and z coordinates of a component, and the fourth input is the distance from that component to the core component in the tree structure. The first three outputs are the probabilities of the modules being a brick, a joint, or empty space, and the last two outputs are the probabilities of the module being rotated 0 or 90 degrees. For both module type and rotation the output with the highest probability is always chosen; randomness is not involved.

The body's genotype to phenotype mapping operates as follows: The core component is generated at the origin. We move outwards from the core component until there are no open sockets(breadth-first exploration), querying the CPPN network to determine the type and rotation of each module. Additionally, we stop when ten modules have been created. The coordinates of each module are integers; a module attached to the front of the core module will have coordinates (0,1,0). If a module would be placed on a location already occupied by a previous module, the module is simply not placed and the branch ends there. In the evolutionary loop for generating the body of offspring, we use the same mutation and crossover operators as in MultiNEAT (\url{https://github.com/MultiNEAT/}).

\subsection*{Robot Controller (Brain)}
\subsubsection*{Brain Phenotype}
We use Central Pattern Generators (CPGs)-based controller to drive the modular robots, which has demonstrated their success in controlling various types of robots, from legged to wheeled ones in previous research. Each joint of the robot has an associated CPG that is defined by three neurons: an $x_i$-neuron, a $y_i$-neuron and an $out_i$-neuron. 
The change of the $x_i$ and $y_i$ neurons' states with respect to time is obtained by multiplying the activation value of the opposite neuron with the corresponding weight  $\dot{x}_i = w_i y_i$, $\dot{y}_i = -w_i x_i$. To reduce the search space we set $w_{x_iy_i}$ to be equal to $-w_{y_ix_i}$ and call their absolute value $w_i$. The resulting activations of neurons $x_i$ and $y_i$ are periodic and bounded. The initial states of all $x$ and $y$ neurons are set to $\frac{\sqrt{2}}{2}$ because this leads to a sine wave with amplitude 1, which matches the limited rotating angle of the joints.


To enable more complex output patterns, connections between CPGs of neighbouring joints are implemented. An example of the CPG network of a "+" shape robot is shown in Figure \ref{fig:cpg_network}. Two joints are said to be neighbours if their distance in the morphology tree is less than or equal to two. 
Consider the $i_{th}$ joint, and $\mathcal{N}_i$ the set of indices of the joints neighbouring it, $w_{ij}$ the weight of the connection between $x_i$ and $x_j$. Again, $w_{ij}$ is set to be $-w_{ji}$. The extended system of differential equations becomes equation \ref{eq:cpg1}.

\begin{minipage}{0.45\textwidth}
    \centering
    \begin{equation}
    \begin{aligned}
        \dot{x}_i &= w_i y_i + \sum_{j \in \mathcal{N}_i} w_{ji} x_j, \hspace{1cm}
        \dot{y}_i &= -w_i x_i
    \end{aligned}
    \label{eq:cpg1}
\end{equation}
\end{minipage}\hfill
\begin{minipage}{0.45\textwidth}
    \centering
    \begin{equation}
        out_{(i,t)}(x_{(i,t)}) = \frac{2}{1+e^{-2x_{(i,t)}}} - 1
    \label{eq:cpg2}
    \end{equation}
\end{minipage}

Because of this addition, $x$ neurons are no longer bounded between $[-1,1]$. For this reason, we use the hyperbolic tangent function (\emph{tanh}) as the activation function of $out_i$-neurons (equation \ref{eq:cpg2}).

\subsubsection*{Brain Genotype}
In biological organisms, including humans, not all genes are actively expressed or used at all times. Gene expression regulation allows cells to control which genes are turned on (expressed) or off (silenced) in response to various internal and external factors. Inspired by this, we utilize a fixed size array-based structure for the brain's genotypic representation to map the CPG weights. It is important to notice that not all the elements of the genotype matrix are going to be used by each robot. This means that their brain's genotype can carry additional information that could be exploited by their children with different morphologies.

The mapping is achieved via direct encoding, a method chosen specifically for its potential to enable reversible encoding in future stages. Every modular robot can be represented as a 3D grid in which the core module occupies the central position and each module's position is given by a triple of coordinates. When building the controller from our genotype, we use the coordinates of the joints in the grid to locate the corresponding CPG weight. To reduce the size of our genotype, instead of the 3D grid, we use a simplified 3D in which the third dimension is removed. For this reason, some joints might end up with the same coordinates and will be dealt with accordingly. 

Since our robots have a maximum of 10 modules, every robot configuration can be represented in a grid of $21 \times 21$. Each joint in a robot can occupy any position of the grid except the center. For this reason, the possible positions of a joint in our morphologies are exactly $(21 \cdot 21) - 1=440$. We can represent all the internal weights of every possible CPG in our morphologies as a $440$-long array. When building the phenotype from this array, we can simply retrieve the corresponding weight starting from a joint's coordinates in the body grid.

To represent the external connections between CPGs, we need to consider all the possible neighbours a joint can have. In the 2-dimensional grid, the number of cells in a distance-2 neighbourhood for each position is represented by the Delannoy number $D(2,2) = 13$, including the central element. Each one of the neighbours can be identified using the relative position from the joint taken into consideration. Since our robots can assume a 3D position, we need to consider an additional connection for modules with the same 2D coordinates.

\begin{figure}

    \begin{minipage}{0.5\textwidth}
    \includegraphics[width=0.99\linewidth]{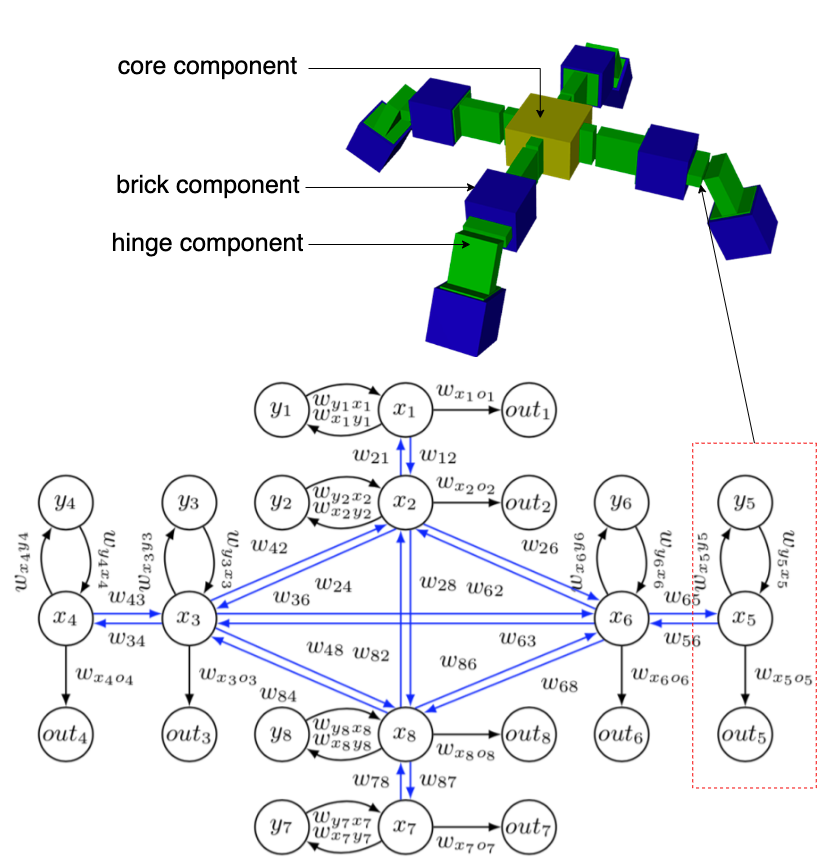}
        \caption{\label{fig:cpg_network}An example of a "+" shape robot and its brain phenotype (CPG network). In our design, the topology of the brain is determined by the topology of the body. The red rectangle is a single CPG which controls a corresponding hinge.}
    \end{minipage}\hfill
    \begin{minipage}{0.45\textwidth}
        \includegraphics[width=0.95\linewidth]{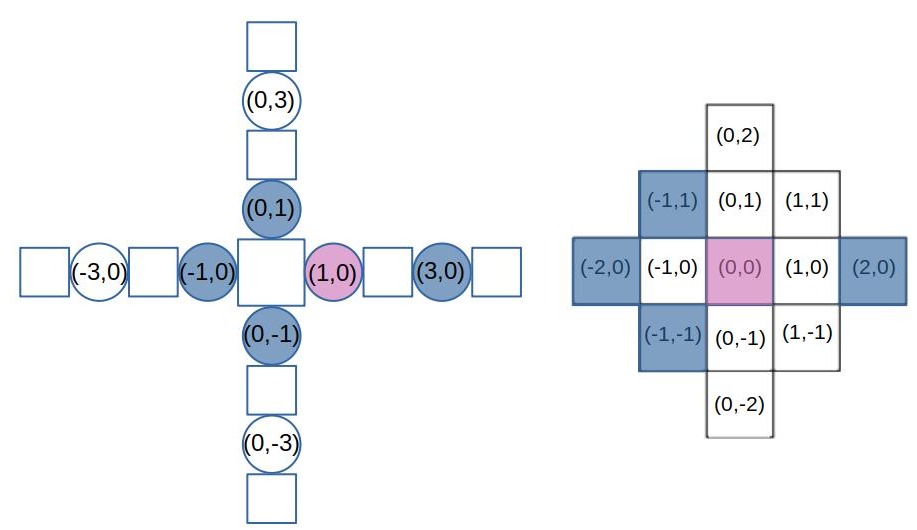}
        \caption{Brain genotype to phenotype mapping of a "+" shape robot. The left image (brain phenotype) shows the schema of the "+" shape robot with the coordinates of its joints in the 2D body grid. The right image (brain genotype) is the distance 2 neighbour of the joint at (1,0). The coordinates reported in the neighbourhood are relative to this joint. The CPG weight of the joint is highlighted in purple and its 2-distance neighbors are in blue.}
        \label{fig:brain_geno}
    \end{minipage}

\end{figure}

To conclude, for each of the $440$ possible joints in the body grid, we need to store 1 internal weight for its CPG, 12 weights for external connections, and 1 weight for connections with CPGs at the same coordinate for a total of 14 weights. The genotype used to represent the robots' brains is an array of size $440 \times 14$. An example of the brain genotype of a "+" shape robot is shown in Figure \ref{fig:brain_geno}.

The recombination operator for the brain genotype is implemented as a uniform crossover where each gene is chosen from either parent with equal probability. The new genotype is generated by essentially flipping a coin for each element of the parents' genotype to decide whether or not it will be included in the offspring's genotype. In the uniform crossover operator, each gene is treated separately.
The mutation operator applies a Gaussian mutation to each element of the genotype by adding a value, with a probability of 0.8, sampled from a Gaussian distribution with 0 mean and 0.5 standard deviation.

\subsection*{Evolution+Learning systems}
\begin{figure*}
    \centering
    \includegraphics[width=0.85\linewidth]{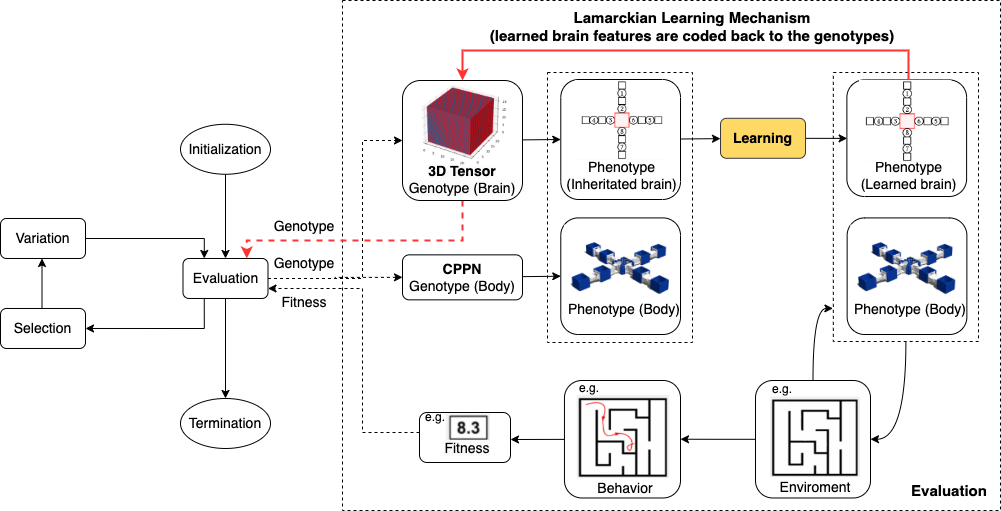}
    \caption{Evolution + Learning framework. This is a general framework for optimizing robots via two interacting adaptive processes. The evolutionary loop (left) optimizes robot morphologies and controllers simultaneously using genotypes that encode both morphologies and controllers. The learning loop (yellow box inside the Evaluation step of the evolutionary loop) optimizes the controller for a given morphology. Note that in general the fitness measure used within the evolutionary loop need not be the same as the quality measure used inside the learning method. With the red lines, it is the Lamarckian learning mechanism which allows the phenotype of the brain coded back to the genotype and pass it to the next generation. Without the red lines, it is the Darwinian learning mechanism. cf. Algorithm \ref{alg:EL}}
    \label{fig:E+L}
\end{figure*}

The complete integrated process of evolution and learning is illustrated in Figure \ref{fig:E+L}, while Algorithm \ref{alg:EL} displays the pseudocode. With the yellow highlighted code, it is the Lamarckian learning mechanism, without it is the Darwinian learning mechanism. Note that for the sake of generality, we distinguish two types of quality testing depending on the context, evolution or learning. Within the evolutionary cycle (line 2 and line 14) a test is called an evaluation and it delivers a fitness value. Inside the learning cycle which is blue highlighted, a test is called an assessment (line 11) and it delivers a reward value. This distinction reflects that in general the notion of fitness can be different from the task performance, perhaps more complex involving more tasks, other behavioral traits not related to any task, or even morphological properties.  

\begin{algorithm}[ht!]
  \caption{Evolution+Learning}
  \label{alg:EL}
  \begin{algorithmic}[1]
    \State INITIALIZE robot population (genotypes + phenotypes with body and brain)  
    \State EVALUATE each robot  (evaluation delivers a fitness value)
    \While{not STOP-EVOLUTION}
        \State SELECT parents; (based on fitness)
        \State RECOMBINE+MUTATE parents' bodies; (this delivers a new body genotype)
        \State RECOMBINE+MUTATE parents' brains; (this delivers a new brain genotype)
        \State CREATE offspring robot body; (this delivers a new body phenotype)
        \State CREATE offspring robot brain; (this delivers a new brain phenotype)
        
        \State INITIALIZE brain(s) for the learning process; (in the new body)

        \tikzmk{A}
        \While{not STOP-LEARNING}
            \State ASSESS offspring; (assessment delivers a reward value)
            \State GENERATE new brain for offspring;
        \EndWhile 
        
        \tikzmk{B} \boxit{blue}
        \State EVALUATE offspring with learned brain; (evaluation delivers a fitness value) 
        
        \tikzmk{A}
        \State UPDATE brain genotype 

        \tikzmk{B} \boxit{yellow}
        \State SELECT survivors / UPDATE population
          
    \EndWhile
 \end{algorithmic}
\end{algorithm}

\vspace{-1em}

\subsubsection*{Evolution loop}
For the outer evolutionary loop, we use a variant of the well-known ($\mu$ + $\lambda$) selection mechanism to update the population. The bodies of the robots are evolved with sexual reproduction while the brains of the robots are evolved with asexual reproduction. 

Body - sexual reproduction: The body of every new offspring is created through recombination and mutation of the genotypes of its parents. Parents are selected from the current generation using binary tournaments with replacement. We perform two tournaments in which two random potential parents are selected. In each tournament the potential parents are compared, the one with the highest fitness wins the tournament and becomes a parent. 

Brain - asexual reproduction: The brain genotype of the best-performing parent is mutated (without recombination) before being inherited by its offspring. This choice is based on preliminary experiments that indicated that asexual brain reproduction is the better method, as it resulted in robots with higher fitness. 

\subsubsection*{Learning loop}

For the inner learning loop which is to search in the space of brain configurations and fine-tune the parameters, we have chosen Reversible Differential Evolution (RevDE) as a learner, because in a recent study on modular robots [\cite{Diggelen2021a}], it was demonstrated that RevDE [\cite{Tomczak2020,weglarz2021population}], an altered version of Differential Evolution, performs and generalizes well across various morphologies. This algorithm works as follows:
\begin{enumerate}
    \item Initialize a population with \textit{$\mu$} samples ($n$-dimensional vectors), $\mathcal{P}_{\mu}$. 
    \item Evaluate all \textit{$\mu$} samples.
    \item Apply the reversible differential mutation operator and the uniform crossover operator.\\
    \textit{The reversible differential mutation operator}: Three new candidates are generated by randomly picking a triplet from the population, $(\mathbf{w}_i,\mathbf{w}_j,\mathbf{w}_k)\in \mathcal{P}_{\mu}$, then all three individuals are perturbed by adding a scaled difference in the following manner:
        \begin{equation}\label{eq:de3}
            \begin{split}
            \mathbf{v}_1 &= \mathbf{w}_i + F \cdot (\mathbf{w}_j-\mathbf{w}_k) \\
            \mathbf{v}_2 &= \mathbf{w}_j + F \cdot (\mathbf{w}_k-\mathbf{v}_1) \\
            \mathbf{v}_3 &= \mathbf{w}_k + F\cdot (\mathbf{v}_1-\mathbf{v}_2) 
            \end{split}
        \end{equation}
        where $F\in R_+$ is the scaling factor. New candidates $y_1$ and $y_2$ are used to calculate perturbations using points outside the population. This approach does not follow the typical construction of an EA where only evaluated candidates are mutated.\\
        \textit{The uniform crossover operator}: Following the original DE method [\cite{Storn1997}], we first sample a binary mask $\mathbf{m} \in \{0, 1\}^D$ according to the Bernoulli distribution with probability \textit{$CR$} shared across $D$ dimensions, and calculate the final candidate according to the following formula:
        \begin{equation}\label{eq:de2}
              \mathbf{u} = \mathbf{m} \odot \mathbf{w}_n+(1-m) \odot \mathbf{w}_n 
        \end{equation}
        Following general recommendations in literature [\cite{Pedersen2010}] to obtain stable exploration behaviour, the crossover probability CR is fixed to a value of $0.9$ and according to the analysis provided in [\cite{Tomczak2020}], the scaling factor $F$ is fixed to a value of 0.5. 
    \item Perform a selection over the population based on the fitness value and select \textit{$\mu$} samples.
    \item Repeat from step (2) until the maximum number of iterations is reached.
\end{enumerate}

As explained above, we apply RevDE here as a learning method for `newborn' robots. In particular, it will be used to optimize the weights of the CPGs of our modular robots for the tasks during the Infancy stage. The initial population of $X = 10$ weight vectors for RevDE is created by using the inherited brain of the given robot. Specifically, the values of the inherited weight vector are altered by adding Gaussian noise to create mutant vectors and the initial population consists of nine such mutants and the vector with the inherited weights.

\subsection*{Task and Fitness function}

Point navigation requires feedback (coordinates)from the environment passing to the controller to steer the robot. The coordinates are used to obtain the angle between the current position and the target. If the target is on the right, the right joints are slowed down and vice versa. 

A robot is spawned at the centre of a flat arena (10 × 10 m2) to reach a sequence of target points $P_1,...,P_N$. In each evaluation, the robot has to reach as many targets in order as possible. Success in this task requires the ability to move fast to reach one target and then quickly change direction to another target in a short duration. A target point is considered to be reached if the robot gets within 0.01 meters from it. To keep runtimes within practically acceptable limits, we set the simulation time per evaluation to be 40 seconds which allows robots to reach at least 2 targets $P_1(1,-1), P_2(0,-2)$.


The data collected from the simulator is the following:
\begin{itemize}
    \item The coordinates of the core component of the robot at the start of the simulation are approximate to $P_0 (0,0)$;
    \item The coordinates of the robot, sampled during the simulation at 5Hz, allowing us to plot and approximate the length of the followed path;
    \item The coordinates of the robot at the end of the simulation $P_T(x_T,y_T)$;
    \item The coordinates of the target points $P_1(x_1,y_1)$... $P_n(x_n,y_n)$.
    \item The coordinates of the robot, sampled during the simulation at 5Hz, allow us to plot and approximate the length of the path $L$.
\end{itemize}

The fitness function is designed to maximize the number of targets reached and to minimize the path length.
\begin{equation} 
    F=\sum_{i=1}^{k}dist(P_i,P_{i-1})
    +(dist(P_k,P_{k-1}) - dist(P_T,P_k))
    - \omega \cdot L
\end{equation}
where $k$ is the number of target points reached by the robot at the end of the evaluation, and $L$ is the path travelled. The first term of the function is a sum of the distances between the target points the robot has reached. The second term is necessary when the robot has not reached all the targets and it calculates the distance travelled toward the next unreached target. The last term is used to penalize longer paths and $\omega$ is a constant scalar that is set to 0.1 in the experiments. E.g., a robot just reached 2 targets, the maximum fitness value will be $dist(P_1,P_0)+(dist(P_2,P_1)-dist(P2,P2))-0.1\cdot L=\sqrt{2}+\sqrt{2}-0.2\cdot\sqrt{2} \approx 2.54$ ($L$ is shortest path length to go through $P_1$ and $P_2$ which is equal to $2\cdot\sqrt{2}$).

\subsection*{Experimental setup}
We use a Mujoco simulator-based wrapper called Revolve2 (https://github.com/ci-group/revolve2) to run experiments. For the ($\mu$ + $\lambda$) selection in the outer evolutionary loop, we set $\mu =50$ and $\lambda = 25$. The evolutionary process is terminated after 30 generations. Therefore, we perform $(25+25\cdot30)$ robots$= 775$ fitness evaluations for each evolutionary loop.

For the inner learning loop, we apply RevDE on each robot body resulting in 280 extra fitness evaluations. This number is based on the learning assessment from RevDE for running 10 initial samples with 10 iterations using $(\mu+\lambda)$ selection. The first iteration contains 10 samples, and from the second iteration onwards each iteration creates 30 new candidates, resulting in a total of $10 + 30 \cdot (10-1)= 280$ evaluations. 

In our research, the fitness measure to drive evolution and the performance measure to drive learning are the same by design. Thus, we use the same test procedure, simulating one robot for 40 simulated seconds for the point navigation task, for the evolutionary as well as the learning trials. 

To sum up, for running the experiments we perform $775\cdot280\cdot2= 434,000$ fitness evaluations which amount to $434,000 \cdot 40/60/60=4,822$ hours of simulated time. To get a robust assessment of the performance all the experiments are repeated 20 times independently. In practice, it takes about 4.25 days to run 2 runs in parallel on 2 64-core processors. The experimental parameters we used in the experiments are described in Table \ref{tab:parameters} in the Supplementary Information section.

\section*{Data availability statement}
The code for replicating this work and carrying out the experiments is available online: \url{https://shorturl.at/epzGS}. 
A short video providing a visual overview of our research is available at \url{https://shorturl.at/ahETW}.

\clearpage
\bibliography{references}

\begin{thebibliography}{10}
\urlstyle{rm}
\expandafter\ifx\csname url\endcsname\relax
  \def\url#1{\texttt{#1}}\fi
\expandafter\ifx\csname urlprefix\endcsname\relax\def\urlprefix{URL }\fi
\expandafter\ifx\csname doiprefix\endcsname\relax\def\doiprefix{DOI: }\fi
\providecommand{\bibinfo}[2]{#2}
\providecommand{\eprint}[2][]{\url{#2}}

\bibitem{nolfi2000evolutionary}
\bibinfo{author}{Nolfi, S.}, \bibinfo{author}{Floreano, D.} \&
  \bibinfo{author}{Floreano, D.~D.}
\newblock \emph{\bibinfo{title}{Evolutionary robotics: The biology,
  intelligence, and technology of self-organizing machines}}
  (\bibinfo{publisher}{MIT press}, \bibinfo{year}{2000}).

\bibitem{doncieux2015evolutionary}
\bibinfo{author}{Doncieux, S.}, \bibinfo{author}{Bredeche, N.},
  \bibinfo{author}{Mouret, J.-B.} \& \bibinfo{author}{Eiben, A.~E.}
\newblock \bibinfo{journal}{\bibinfo{title}{Evolutionary robotics: what, why,
  and where to}}.
\newblock {\emph{\JournalTitle{Frontiers in Robotics and AI}}}
  \textbf{\bibinfo{volume}{2}}, \bibinfo{pages}{4} (\bibinfo{year}{2015}).

\bibitem{Floreano1998}
\bibinfo{author}{Floreano, D.} \& \bibinfo{author}{Mondada, F.}
\newblock \bibinfo{journal}{\bibinfo{title}{Evolutionary neurocontrollers for
  autonomous mobile robots}}.
\newblock {\emph{\JournalTitle{Neural networks}}}
  \textbf{\bibinfo{volume}{11}}, \bibinfo{pages}{1461--1478}
  (\bibinfo{year}{1998}).

\bibitem{Capi2008}
\bibinfo{author}{Capi, G.}, \bibinfo{author}{Pojani, G.} \&
  \bibinfo{author}{Kaneko, S.-I.}
\newblock \bibinfo{title}{Evolution of task switching behaviors in real mobile
  robots}.
\newblock In \emph{\bibinfo{booktitle}{2008 3rd International Conference on
  Innovative Computing Information and Control}}, \bibinfo{pages}{495--495},
  \doiprefix\url{10.1109/ICICIC.2008.261} (\bibinfo{year}{2008}).

\bibitem{sims1994evolving}
\bibinfo{author}{Sims, K.}
\newblock \bibinfo{title}{Evolving virtual creatures}.
\newblock In \emph{\bibinfo{booktitle}{Proceedings of the 21st annual
  conference on Computer graphics and interactive techniques}},
  \bibinfo{pages}{15--22} (\bibinfo{organization}{ACM}, \bibinfo{year}{1994}).

\bibitem{auerbach2012relationship}
\bibinfo{author}{Auerbach, J.~E.} \& \bibinfo{author}{Bongard, J.~C.}
\newblock \bibinfo{title}{On the relationship between environmental and
  morphological complexity in evolved robots}.
\newblock In \emph{\bibinfo{booktitle}{Proceedings of the 14th annual
  conference on Genetic and evolutionary computation}},
  \bibinfo{pages}{521--528} (\bibinfo{organization}{ACM},
  \bibinfo{year}{2012}).

\bibitem{Weel2014}
\bibinfo{author}{Weel, B.}, \bibinfo{author}{Crosato, E.},
  \bibinfo{author}{Heinerman, J.}, \bibinfo{author}{Haasdijk, E.} \&
  \bibinfo{author}{Eiben, A.~E.}
\newblock \bibinfo{title}{{A Robotic Ecosystem with Evolvable Minds and
  Bodies}}.
\newblock In \emph{\bibinfo{booktitle}{2014 IEEE International Conference on
  Evolvable Systems}}, \bibinfo{pages}{165 -- 172} (\bibinfo{publisher}{IEEE
  Press, Piscataway, NJ}, \bibinfo{year}{2014}).

\bibitem{Lipson2016}
\bibinfo{author}{Lipson, H.}, \bibinfo{author}{SunSpiral, V.},
  \bibinfo{author}{Bongard, J.~C.} \& \bibinfo{author}{Cheney, N.}
\newblock \bibinfo{journal}{\bibinfo{title}{On the difficulty of co-optimizing
  morphology and control in evolved virtual creatures}}.
\newblock {\emph{\JournalTitle{Artificial Life}}} \bibinfo{pages}{226--233}
  (\bibinfo{year}{2016}).

\bibitem{Cheney2018}
\bibinfo{author}{Cheney, N.}, \bibinfo{author}{Bongard, J.},
  \bibinfo{author}{SunSpiral, V.} \& \bibinfo{author}{Lipson, H.}
\newblock \bibinfo{journal}{\bibinfo{title}{{Scalable co-optimization of
  morphology and control in embodied machines}}}.
\newblock {\emph{\JournalTitle{Journal of the Royal Society Interface}}}
  \textbf{\bibinfo{volume}{15}} (\bibinfo{year}{2018}).

\bibitem{Miras-Eiben-2019}
\bibinfo{author}{Miras, K.} \& \bibinfo{author}{Eiben, A.~E.}
\newblock \bibinfo{title}{Effects of environmental conditions on evolved robot
  morphologies and behavior}.
\newblock In \emph{\bibinfo{booktitle}{Proceedings of the Genetic and
  Evolutionary Computation Conference}}, GECCO '19, \bibinfo{pages}{125–132},
  \doiprefix\url{10.1145/3321707.3321811} (\bibinfo{publisher}{Association for
  Computing Machinery}, \bibinfo{address}{New York, NY, USA},
  \bibinfo{year}{2019}).

\bibitem{Hockings2020}
\bibinfo{author}{Hockings, N.} \& \bibinfo{author}{Howard, D.}
\newblock \bibinfo{journal}{\bibinfo{title}{New biological morphogenetic
  methods for evolutionary design of robot bodies}}.
\newblock {\emph{\JournalTitle{Frontiers in Bioengineering and Biotechnology}}}
  \textbf{\bibinfo{volume}{8}}, \doiprefix\url{10.3389/fbioe.2020.00621}
  (\bibinfo{year}{2020}).

\bibitem{Stensby2021}
\bibinfo{author}{Stensby, E.~H.}, \bibinfo{author}{Ellefsen, K.~O.} \&
  \bibinfo{author}{Glette, K.}
\newblock \bibinfo{title}{Co-optimising robot morphology and controller in a
  simulated open-ended environment}.
\newblock In \emph{\bibinfo{booktitle}{Applications of Evolutionary
  Computation: 24th International Conference, EvoApplications 2021, Held as
  Part of EvoStar 2021, Virtual Event, April 7–9, 2021, Proceedings}},
  \bibinfo{pages}{34–49}, \doiprefix\url{10.1007/978-3-030-72699-7_3}
  (\bibinfo{publisher}{Springer-Verlag}, \bibinfo{address}{Berlin, Heidelberg},
  \bibinfo{year}{2021}).

\bibitem{Medvet2021}
\bibinfo{author}{Medvet, E.}, \bibinfo{author}{Bartoli, A.},
  \bibinfo{author}{Pigozzi, F.} \& \bibinfo{author}{Rochelli, M.}
\newblock \bibinfo{journal}{\bibinfo{title}{{Biodiversity in evolved
  voxel-based soft robots}}}.
\newblock {\emph{\JournalTitle{Proceedings of the 2021 Genetic and Evolutionary
  Computation Conference}}} \bibinfo{pages}{129--137} (\bibinfo{year}{2021}).

\bibitem{Nygaard2021}
\bibinfo{author}{Nygaard, T.~F.}, \bibinfo{author}{Martin, C.~P.},
  \bibinfo{author}{Torresen, J.}, \bibinfo{author}{Glette, K.} \&
  \bibinfo{author}{Howard, D.}
\newblock \bibinfo{journal}{\bibinfo{title}{Real-world embodied ai through a
  morphologically adaptive quadruped robot}}.
\newblock {\emph{\JournalTitle{Nature Machine Intelligence}}}
  \textbf{\bibinfo{volume}{1}}, \bibinfo{pages}{12--19} (\bibinfo{year}{2021}).

\bibitem{Nolfi1999}
\bibinfo{author}{Nolfi, S.} \& \bibinfo{author}{Floreano, D.}
\newblock \bibinfo{journal}{\bibinfo{title}{Learning and evolution}}.
\newblock {\emph{\JournalTitle{Autonomous robots}}}
  \textbf{\bibinfo{volume}{7}}, \bibinfo{pages}{89--113}
  (\bibinfo{year}{1999}).

\bibitem{Floreano1996}
\bibinfo{author}{Floreano, D.} \& \bibinfo{author}{Mondada, F.}
\newblock \bibinfo{title}{Evolution of plastic neurocontrollers for situated
  agents}.
\newblock In \emph{\bibinfo{booktitle}{Proc. of The Fourth International
  Conference on Simulation of Adaptive Behavior (SAB), From Animals to
  Animats}} (\bibinfo{organization}{ETH Z{\"u}rich}, \bibinfo{year}{1996}).

\bibitem{Harvey1997}
\bibinfo{author}{Harvey, I.}, \bibinfo{author}{Husbands, P.},
  \bibinfo{author}{Cliff, D.}, \bibinfo{author}{Thompson, A.} \&
  \bibinfo{author}{Jakobi, N.}
\newblock \bibinfo{journal}{\bibinfo{title}{Evolutionary robotics: the sussex
  approach}}.
\newblock {\emph{\JournalTitle{Robotics and autonomous systems}}}
  \textbf{\bibinfo{volume}{20}}, \bibinfo{pages}{205--224}
  (\bibinfo{year}{1997}).

\bibitem{Schembri2007}
\bibinfo{author}{Schembri, M.}, \bibinfo{author}{Mirolli, M.} \&
  \bibinfo{author}{Baldassarre, G.}
\newblock \bibinfo{title}{Evolution and learning in an intrinsically motivated
  reinforcement learning robot}.
\newblock In \emph{\bibinfo{booktitle}{Proceedings of the 9th European
  Conference on Advances in Artificial Life}}, vol. \bibinfo{volume}{4648},
  \bibinfo{pages}{294--303} (\bibinfo{year}{2007}).

\bibitem{Sproewitz2008}
\bibinfo{author}{Sproewitz, A.}, \bibinfo{author}{Moeckel, R.},
  \bibinfo{author}{Maye, J.} \& \bibinfo{author}{Ijspeert, A.~J.}
\newblock \bibinfo{journal}{\bibinfo{title}{Learning to move in modular robots
  using central pattern generators and online optimization}}.
\newblock {\emph{\JournalTitle{The International Journal of Robotics
  Research}}} \textbf{\bibinfo{volume}{27}}, \bibinfo{pages}{423--443}
  (\bibinfo{year}{2008}).

\bibitem{Bellas2010}
\bibinfo{author}{Bellas, F.}, \bibinfo{author}{Duro, R.~J.},
  \bibinfo{author}{Fai{\~n}a, A.} \& \bibinfo{author}{Souto, D.}
\newblock \bibinfo{journal}{\bibinfo{title}{Multilevel darwinist brain (mdb):
  Artificial evolution in a cognitive architecture for real robots}}.
\newblock {\emph{\JournalTitle{IEEE Transactions on autonomous mental
  development}}} \textbf{\bibinfo{volume}{2}}, \bibinfo{pages}{340--354}
  (\bibinfo{year}{2010}).

\bibitem{Khazanov2014}
\bibinfo{author}{Khazanov, M.}, \bibinfo{author}{Jocque, J.} \&
  \bibinfo{author}{Rieffel, J.}
\newblock \bibinfo{title}{Evolution of locomotion on a physical tensegrity
  robot}.
\newblock In \emph{\bibinfo{booktitle}{ALIFE 14: The Fourteenth International
  Conference on the Synthesis and Simulation of Living Systems}},
  \bibinfo{pages}{232--238} (\bibinfo{organization}{MIT Press},
  \bibinfo{year}{2014}).

\bibitem{Ruud2017}
\bibinfo{author}{Ruud, E.~L.}, \bibinfo{author}{Samuelsen, E.} \&
  \bibinfo{author}{Glette, K.}
\newblock \bibinfo{journal}{\bibinfo{title}{{Memetic robot control evolution
  and adaption to reality}}}.
\newblock {\emph{\JournalTitle{2016 IEEE Symposium Series on Computational
  Intelligence, SSCI 2016}}}  (\bibinfo{year}{2017}).

\bibitem{Schaff2019}
\bibinfo{author}{Schaff, C.}, \bibinfo{author}{Yunis, D.},
  \bibinfo{author}{Chakrabarti, A.} \& \bibinfo{author}{Walter, M.~R.}
\newblock \bibinfo{title}{Jointly learning to construct and control agents
  using deep reinforcement learning}.
\newblock In \emph{\bibinfo{booktitle}{2019 International Conference on
  Robotics and Automation (ICRA)}}, \bibinfo{pages}{9798--9805}
  (\bibinfo{organization}{IEEE}, \bibinfo{year}{2019}).

\bibitem{Hwangbo2019}
\bibinfo{author}{Hwangbo, J.} \emph{et~al.}
\newblock \bibinfo{journal}{\bibinfo{title}{Learning agile and dynamic motor
  skills for legged robots}}.
\newblock {\emph{\JournalTitle{Science Robotics}}}
  \textbf{\bibinfo{volume}{4}}, \bibinfo{pages}{eaau5872}
  (\bibinfo{year}{2019}).

\bibitem{LeGoff2020}
\bibinfo{author}{{Le Goff}, L.~K.} \emph{et~al.}
\newblock \bibinfo{title}{{Sample and time efficient policy learning with
  CMA-ES and Bayesian Optimisation}}.
\newblock In \emph{\bibinfo{booktitle}{The 2020 Conference on Artificial
  Life}}, \bibinfo{number}{January}, \bibinfo{pages}{2020}
  (\bibinfo{year}{2020}).

\bibitem{Lehman2011}
\bibinfo{author}{Lehman, J.} \& \bibinfo{author}{Stanley, K.~O.}
\newblock \bibinfo{title}{Evolving a diversity of virtual creatures through
  novelty search and local competition}.
\newblock In \emph{\bibinfo{booktitle}{Proceedings of the 13th annual
  conference on Genetic and evolutionary computation}},
  \bibinfo{pages}{211--218} (\bibinfo{year}{2011}).

\bibitem{Jelisavcic2017}
\bibinfo{author}{Jelisavcic, M.} \emph{et~al.}
\newblock \bibinfo{journal}{\bibinfo{title}{Real-world evolution of robot
  morphologies: A proof of concept}}.
\newblock {\emph{\JournalTitle{Artificial life}}}
  \textbf{\bibinfo{volume}{23}}, \bibinfo{pages}{206--235}
  (\bibinfo{year}{2017}).

\bibitem{Kriegman2018}
\bibinfo{author}{Kriegman, S.}, \bibinfo{author}{Cheney, N.} \&
  \bibinfo{author}{Bongard, J.}
\newblock \bibinfo{journal}{\bibinfo{title}{How morphological development can
  guide evolution}}.
\newblock {\emph{\JournalTitle{Scientific reports}}}
  \textbf{\bibinfo{volume}{8}}, \bibinfo{pages}{1--10} (\bibinfo{year}{2018}).

\bibitem{Aguilar2019}
\bibinfo{author}{Aguilar, L.}, \bibinfo{author}{Bennati, S.} \&
  \bibinfo{author}{Helbing, D.}
\newblock \bibinfo{journal}{\bibinfo{title}{How learning can change the course
  of evolution}}.
\newblock {\emph{\JournalTitle{Plos one}}} \textbf{\bibinfo{volume}{14}},
  \bibinfo{pages}{e0219502} (\bibinfo{year}{2019}).

\bibitem{Howard2019}
\bibinfo{author}{Howard, D.} \emph{et~al.}
\newblock \bibinfo{journal}{\bibinfo{title}{Evolving embodied intelligence from
  materials to machines}}.
\newblock {\emph{\JournalTitle{Nature Machine Intelligence}}}
  \textbf{\bibinfo{volume}{1}}, \bibinfo{pages}{12--19} (\bibinfo{year}{2019}).

\bibitem{Miras2020}
\bibinfo{author}{Miras, K.}, \bibinfo{author}{De~Carlo, M.},
  \bibinfo{author}{Akhatou, S.} \& \bibinfo{author}{Eiben, A.~E.}
\newblock \bibinfo{title}{Evolving-controllers versus learning-controllers for
  morphologically evolvable robots}.
\newblock In \emph{\bibinfo{booktitle}{Applications of Evolutionary
  Computation}}, \bibinfo{pages}{86--99} (\bibinfo{year}{2020}).

\bibitem{Gupta2021}
\bibinfo{author}{Gupta, A.}, \bibinfo{author}{Savarese, S.},
  \bibinfo{author}{Ganguli, S.} \& \bibinfo{author}{Fei-Fei, L.}
\newblock \bibinfo{journal}{\bibinfo{title}{{Embodied Intelligence via Learning
  and Evolution}}}.
\newblock {\emph{\JournalTitle{Nature Communications}}}
  \textbf{\bibinfo{volume}{12}} (\bibinfo{year}{2021}).

\bibitem{Goff2021a}
\bibinfo{author}{Goff, L. K.~L.} \& \bibinfo{author}{Hart, E.}
\newblock \bibinfo{title}{On the challenges of jointly optimising robot
  morphology and control using a hierarchical optimisation scheme}.
\newblock In \emph{\bibinfo{booktitle}{Proceedings of the Genetic and
  Evolutionary Computation Conference Companion}}, \bibinfo{pages}{1498--1502}
  (\bibinfo{year}{2021}).

\bibitem{Hart2022}
\bibinfo{author}{Hart, E.} \& \bibinfo{author}{Le~Goff, L.~K.}
\newblock \bibinfo{journal}{\bibinfo{title}{Artificial evolution of robot
  bodies and control: on the interaction between evolution, learning and
  culture}}.
\newblock {\emph{\JournalTitle{Philosophical Transactions of the Royal Society
  B}}} \textbf{\bibinfo{volume}{377}}, \bibinfo{pages}{20210117}
  (\bibinfo{year}{2022}).

\bibitem{Luo2022}
\bibinfo{author}{Luo, J.}, \bibinfo{author}{Stuurman, A.},
  \bibinfo{author}{Tomczak, J.~M.}, \bibinfo{author}{Ellers, J.} \&
  \bibinfo{author}{Eiben, A.~E.}
\newblock \bibinfo{journal}{\bibinfo{title}{The effects of learning in
  morphologically evolving robot systems}}.
\newblock {\emph{\JournalTitle{Frontiers in Robotics and AI}}}
  \textbf{\bibinfo{volume}{5}} (\bibinfo{year}{2022}).

\bibitem{wiser2013long}
\bibinfo{author}{Wiser, M.~J.}, \bibinfo{author}{Ribeck, N.} \&
  \bibinfo{author}{Lenski, R.~E.}
\newblock \bibinfo{journal}{\bibinfo{title}{Long-term dynamics of adaptation in
  asexual populations}}.
\newblock {\emph{\JournalTitle{Science}}} \textbf{\bibinfo{volume}{342}},
  \bibinfo{pages}{1364--1367} (\bibinfo{year}{2013}).

\bibitem{long2012darwin}
\bibinfo{author}{Long, J.}
\newblock \emph{\bibinfo{title}{Darwin's devices: What evolving robots can
  teach us about the history of life and the future of technology}}
  (\bibinfo{publisher}{Basic Books (AZ)}, \bibinfo{year}{2012}).

\bibitem{maynard1992byte}
\bibinfo{author}{Maynard~Smith, J.}
\newblock \bibinfo{journal}{\bibinfo{title}{Byte-sized evolution}}.
\newblock {\emph{\JournalTitle{Nature}}} \textbf{\bibinfo{volume}{355}},
  \bibinfo{pages}{772--773} (\bibinfo{year}{1992}).

\bibitem{montanier2011surviving}
\bibinfo{author}{Montanier, J.-M.} \& \bibinfo{author}{Bredeche, N.}
\newblock \bibinfo{title}{Surviving the tragedy of commons: emergence of
  altruism in a population of evolving autonomous agents}.
\newblock In \emph{\bibinfo{booktitle}{European Conference on Artificial Life}}
  (\bibinfo{year}{2011}).

\bibitem{waibel2011quantitative}
\bibinfo{author}{Waibel, M.}, \bibinfo{author}{Floreano, D.} \&
  \bibinfo{author}{Keller, L.}
\newblock \bibinfo{journal}{\bibinfo{title}{A quantitative test of hamilton's
  rule for the evolution of altruism}}.
\newblock {\emph{\JournalTitle{PLoS biology}}} \textbf{\bibinfo{volume}{9}},
  \bibinfo{pages}{e1000615} (\bibinfo{year}{2011}).

\bibitem{solomon2012comparison}
\bibinfo{author}{Solomon, M.}, \bibinfo{author}{Soule, T.} \&
  \bibinfo{author}{Heckendorn, R.~B.}
\newblock \bibinfo{title}{A comparison of a communication strategies in
  cooperative learning}.
\newblock In \emph{\bibinfo{booktitle}{Proceedings of the 14th annual
  conference on Genetic and evolutionary computation}},
  \bibinfo{pages}{153--160} (\bibinfo{year}{2012}).

\bibitem{floreano2007evolutionary}
\bibinfo{author}{Floreano, D.}, \bibinfo{author}{Mitri, S.},
  \bibinfo{author}{Magnenat, S.} \& \bibinfo{author}{Keller, L.}
\newblock \bibinfo{journal}{\bibinfo{title}{Evolutionary conditions for the
  emergence of communication in robots}}.
\newblock {\emph{\JournalTitle{Current biology}}}
  \textbf{\bibinfo{volume}{17}}, \bibinfo{pages}{514--519}
  (\bibinfo{year}{2007}).

\bibitem{mitri2009evolution}
\bibinfo{author}{Mitri, S.}, \bibinfo{author}{Floreano, D.} \&
  \bibinfo{author}{Keller, L.}
\newblock \bibinfo{journal}{\bibinfo{title}{The evolution of information
  suppression in communicating robots with conflicting interests}}.
\newblock {\emph{\JournalTitle{Proceedings of the National Academy of
  Sciences}}} \textbf{\bibinfo{volume}{106}}, \bibinfo{pages}{15786--15790}
  (\bibinfo{year}{2009}).

\bibitem{wischmann2012historical}
\bibinfo{author}{Wischmann, S.}, \bibinfo{author}{Floreano, D.} \&
  \bibinfo{author}{Keller, L.}
\newblock \bibinfo{journal}{\bibinfo{title}{Historical contingency affects
  signaling strategies and competitive abilities in evolving populations of
  simulated robots}}.
\newblock {\emph{\JournalTitle{Proceedings of the National Academy of
  Sciences}}} \textbf{\bibinfo{volume}{109}}, \bibinfo{pages}{864--868}
  (\bibinfo{year}{2012}).

\bibitem{bongard2011morphological}
\bibinfo{author}{Bongard, J.}
\newblock \bibinfo{journal}{\bibinfo{title}{Morphological change in machines
  accelerates the evolution of robust behavior}}.
\newblock {\emph{\JournalTitle{Proceedings of the National Academy of
  Sciences}}} \textbf{\bibinfo{volume}{108}}, \bibinfo{pages}{1234--1239}
  (\bibinfo{year}{2011}).

\bibitem{auerbach2014environmental}
\bibinfo{author}{Auerbach, J.~E.} \& \bibinfo{author}{Bongard, J.~C.}
\newblock \bibinfo{journal}{\bibinfo{title}{Environmental influence on the
  evolution of morphological complexity in machines}}.
\newblock {\emph{\JournalTitle{PLoS computational biology}}}
  \textbf{\bibinfo{volume}{10}}, \bibinfo{pages}{e1003399}
  (\bibinfo{year}{2014}).

\bibitem{olson2013predator}
\bibinfo{author}{Olson, R.~S.}, \bibinfo{author}{Hintze, A.},
  \bibinfo{author}{Dyer, F.~C.}, \bibinfo{author}{Knoester, D.~B.} \&
  \bibinfo{author}{Adami, C.}
\newblock \bibinfo{journal}{\bibinfo{title}{Predator confusion is sufficient to
  evolve swarming behaviour}}.
\newblock {\emph{\JournalTitle{Journal of The Royal Society Interface}}}
  \textbf{\bibinfo{volume}{10}}, \bibinfo{pages}{20130305}
  (\bibinfo{year}{2013}).

\bibitem{Conway1946}
\bibinfo{author}{Zirkle, C.}
\newblock \bibinfo{journal}{\bibinfo{title}{The early history of the idea of
  the inheritance of acquired characters and of pangenesis}}.
\newblock {\emph{\JournalTitle{Transactions of the American Philosophical
  Society}}} \textbf{\bibinfo{volume}{35}}, \bibinfo{pages}{91--151}
  (\bibinfo{year}{1946}).

\bibitem{Burkhardt2013}
\bibinfo{author}{Burkhardt, R.~W.}
\newblock \bibinfo{title}{Lamarck, evolution, and the inheritance of acquired
  characters}.
\newblock In \emph{\bibinfo{booktitle}{Genetics}}, \bibinfo{pages}{793–805},
  \doiprefix\url{https://doi.org/10.1534/genetics.113.151852}
  (\bibinfo{year}{2013}).

\bibitem{bossdorf2008epigenetics}
\bibinfo{author}{Bossdorf, O.}, \bibinfo{author}{Richards, C.~L.} \&
  \bibinfo{author}{Pigliucci, M.}
\newblock \bibinfo{journal}{\bibinfo{title}{Epigenetics for ecologists}}.
\newblock {\emph{\JournalTitle{Ecology letters}}}
  \textbf{\bibinfo{volume}{11}}, \bibinfo{pages}{106--115}
  (\bibinfo{year}{2008}).

\bibitem{Goff2021}
\bibinfo{author}{Goff, L. K.~L.} \emph{et~al.}
\newblock \bibinfo{journal}{\bibinfo{title}{{Morpho-evolution with learning
  using a controller archive as an inheritance mechanism}}}.
\newblock {\emph{\JournalTitle{http://arxiv.org/abs/2104.04269}}}
  (\bibinfo{year}{2021}).

\bibitem{mingo2013}
\bibinfo{author}{Mingo, J.~M.}, \bibinfo{author}{Aler, R.},
  \bibinfo{author}{Maravall, D.} \& \bibinfo{author}{de~Lope, J.}
\newblock \bibinfo{journal}{\bibinfo{title}{Investigations into lamarckism,
  baldwinism and local search in grammatical evolution guided by
  reinforcement}}.
\newblock {\emph{\JournalTitle{Computing and Informatics}}}
  \textbf{\bibinfo{volume}{32}}, \bibinfo{pages}{595--627}
  (\bibinfo{year}{2013}).

\bibitem{whitley1994}
\bibinfo{author}{Whitley, D.}, \bibinfo{author}{Gordon, V.~S.} \&
  \bibinfo{author}{Mathias, K.}
\newblock \bibinfo{title}{Lamarckian evolution, the baldwin effect and function
  optimization}.
\newblock In \emph{\bibinfo{booktitle}{Parallel Problem Solving from
  Nature—PPSN III: International Conference on Evolutionary Computation The
  Third Conference on Parallel Problem Solving from Nature Jerusalem, Israel,
  October 9--14, 1994 Proceedings 3}}, \bibinfo{pages}{5--15}
  (\bibinfo{organization}{Springer}, \bibinfo{year}{1994}).

\bibitem{Holzinger2014}
\bibinfo{author}{Holzinger, A.} \emph{et~al.}
\newblock \bibinfo{title}{Darwin, lamarck, or baldwin: Applying evolutionary
  algorithms to machine learning techniques}.
\newblock In \emph{\bibinfo{booktitle}{2014 IEEE/WIC/ACM International Joint
  Conferences on Web Intelligence (WI) and Intelligent Agent Technologies
  (IAT)}}, vol.~\bibinfo{volume}{2}, \bibinfo{pages}{449--453},
  \doiprefix\url{10.1109/WI-IAT.2014.132} (\bibinfo{year}{2014}).

\bibitem{mikami1996}
\bibinfo{author}{Mikami, S.}, \bibinfo{author}{Wada, M.} \&
  \bibinfo{author}{Kakazu, Y.}
\newblock \bibinfo{title}{Combining reinforcement learning with ga to find
  co-ordinated control rules for multi-agent system}.
\newblock In \emph{\bibinfo{booktitle}{Proceedings of IEEE International
  Conference on Evolutionary Computation}}, \bibinfo{pages}{356--361}
  (\bibinfo{organization}{IEEE}, \bibinfo{year}{1996}).

\bibitem{ku1999}
\bibinfo{author}{Ku, K. W.~C.}, \bibinfo{author}{Mak, M.~W.} \&
  \bibinfo{author}{Siu, W.~C.}
\newblock \bibinfo{journal}{\bibinfo{title}{Adding learning to cellular genetic
  algorithms for training recurrent neural networks}}.
\newblock {\emph{\JournalTitle{IEEE Transactions on Neural Networks}}}
  \textbf{\bibinfo{volume}{10}}, \bibinfo{pages}{239--252}
  (\bibinfo{year}{1999}).

\bibitem{houck1997}
\bibinfo{author}{Houck, C.~R.}, \bibinfo{author}{Joines, J.~A.},
  \bibinfo{author}{Kay, M.~G.} \& \bibinfo{author}{Wilson, J.~R.}
\newblock \bibinfo{journal}{\bibinfo{title}{Empirical investigation of the
  benefits of partial lamarckianism}}.
\newblock {\emph{\JournalTitle{Evolutionary Computation}}}
  \textbf{\bibinfo{volume}{5}}, \bibinfo{pages}{31--60} (\bibinfo{year}{1997}).

\bibitem{castillo2006}
\bibinfo{author}{Castillo, P.~A.} \emph{et~al.}
\newblock \bibinfo{title}{Lamarckian evolution and the baldwin effect in
  evolutionary neural networks} (\bibinfo{year}{2006}).
\newblock \eprint{cs/0603004}.

\bibitem{ZHANG2013}
\bibinfo{author}{Zhang, C.}, \bibinfo{author}{Chen, J.} \&
  \bibinfo{author}{Xin, B.}
\newblock \bibinfo{journal}{\bibinfo{title}{Distributed memetic differential
  evolution with the synergy of lamarckian and baldwinian learning}}.
\newblock {\emph{\JournalTitle{Applied Soft Computing}}}
  \textbf{\bibinfo{volume}{13}}, \bibinfo{pages}{2947--2959},
  \doiprefix\url{https://doi.org/10.1016/j.asoc.2012.02.028}
  (\bibinfo{year}{2013}).

\bibitem{elsken2018}
\bibinfo{author}{Elsken, T.}, \bibinfo{author}{Metzen, J.~H.} \&
  \bibinfo{author}{Hutter, F.}
\newblock \bibinfo{journal}{\bibinfo{title}{Efficient multi-objective neural
  architecture search via lamarckian evolution}}.
\newblock {\emph{\JournalTitle{arXiv preprint arXiv:1804.09081}}}
  (\bibinfo{year}{2018}).

\bibitem{Nishiwaki2015}
\bibinfo{author}{Nishiwaki, Y.}, \bibinfo{author}{Mukosaka, N.},
  \bibinfo{author}{Tanev, I.} \& \bibinfo{author}{Shimohara, K.}
\newblock \bibinfo{journal}{\bibinfo{title}{On the effects of epigenetic
  programming on the efficiency of incremental evolution of the simulated
  khepera robot}}.
\newblock {\emph{\JournalTitle{Journal of Robotics, Networking and Artificial
  Life}}} \textbf{\bibinfo{volume}{2}}, \bibinfo{pages}{108},
  \doiprefix\url{10.2991/jrnal.2015.2.2.9} (\bibinfo{year}{2015}).

\bibitem{Grefenstette1991}
\bibinfo{author}{Grefenstette, J.~J.}
\newblock \bibinfo{title}{Lamarckian learning in multi-agent environments}.
\newblock In \emph{\bibinfo{booktitle}{International Conference on Genetic
  Algorithms}} (\bibinfo{year}{1991}).

\bibitem{jelisavcic2019lamarckian}
\bibinfo{author}{Jelisavcic, M.}, \bibinfo{author}{Glette, K.},
  \bibinfo{author}{Haasdijk, E.} \& \bibinfo{author}{Eiben, A.}
\newblock \bibinfo{journal}{\bibinfo{title}{Lamarckian evolution of simulated
  modular robots}}.
\newblock {\emph{\JournalTitle{Frontiers in Robotics and AI}}}
  \textbf{\bibinfo{volume}{6}}, \bibinfo{pages}{9} (\bibinfo{year}{2019}).

\bibitem{jelisavcic2018morphological}
\bibinfo{author}{Jelisavcic, M.}, \bibinfo{author}{Miras, K.} \&
  \bibinfo{author}{Eiben, A.}
\newblock \bibinfo{title}{Morphological attractors in darwinian and lamarckian
  evolutionary robot systems}.
\newblock In \emph{\bibinfo{booktitle}{2018 IEEE Symposium Series on
  Computational Intelligence (SSCI)}}, \bibinfo{pages}{859--866}
  (\bibinfo{organization}{IEEE}, \bibinfo{year}{2018}).

\bibitem{miras2018search}
\bibinfo{author}{Miras, K.}, \bibinfo{author}{Haasdijk, E.},
  \bibinfo{author}{Glette, K.} \& \bibinfo{author}{Eiben, A.}
\newblock \bibinfo{title}{Search space analysis of evolvable robot
  morphologies}.
\newblock In \emph{\bibinfo{booktitle}{International Conference on the
  Applications of Evolutionary Computation}}, \bibinfo{pages}{703--718}
  (\bibinfo{organization}{Springer}, \bibinfo{year}{2018}).

\bibitem{Pawlik2015}
\bibinfo{author}{Pawlik, M.} \& \bibinfo{author}{Augsten, N.}
\newblock \bibinfo{journal}{\bibinfo{title}{Tree edit distance: Robust and
  memory-efficient}}.
\newblock {\emph{\JournalTitle{Information Systems}}}
  \textbf{\bibinfo{volume}{56}}, \doiprefix\url{10.1016/j.is.2015.08.004}
  (\bibinfo{year}{2015}).

\bibitem{Auerbach2014}
\bibinfo{author}{Auerbach, J.~E.} \emph{et~al.}
\newblock \bibinfo{title}{{Robogen: Robot generation through artificial
  evolution}}.
\newblock In \emph{\bibinfo{booktitle}{Proceedings of the 14th International
  Conference on the Synthesis and Simulation of Living Systems, ALIFE 2014}},
  \bibinfo{pages}{136--137} (\bibinfo{year}{2014}).

\bibitem{Stanley2007}
\bibinfo{author}{Stanley, K.~O.}
\newblock \bibinfo{journal}{\bibinfo{title}{{Compositional pattern producing
  networks: A novel abstraction of development}}}.
\newblock {\emph{\JournalTitle{Genetic Programming and Evolvable Machines}}}
  \textbf{\bibinfo{volume}{8}}, \bibinfo{pages}{131--162}
  (\bibinfo{year}{2007}).

\bibitem{Diggelen2021a}
\bibinfo{author}{van Diggelen, F.}, \bibinfo{author}{Ferrante, E.} \&
  \bibinfo{author}{Eiben, A.~E.}
\newblock \bibinfo{title}{Comparing lifetime learning methods for
  morphologically evolving robots}.
\newblock In \emph{\bibinfo{booktitle}{GECCO '21: Proceedings of the Genetic
  and Evolutionary Computation Conference Companion}}, \bibinfo{pages}{93--94}
  (\bibinfo{year}{2021}).

\bibitem{Tomczak2020}
\bibinfo{author}{Tomczak, J.~M.}, \bibinfo{author}{Weglarz-Tomczak, E.} \&
  \bibinfo{author}{Eiben, A.~E.}
\newblock \bibinfo{title}{{Differential Evolution with Reversible Linear
  Transformations}}.
\newblock In \emph{\bibinfo{booktitle}{Proceedings of the 2020 Genetic and
  Evolutionary Computation Conference Companion}}, \bibinfo{pages}{205--206}
  (\bibinfo{year}{2020}).
\newblock \eprint{2002.02869}.

\bibitem{weglarz2021population}
\bibinfo{author}{Weglarz-Tomczak, E.}, \bibinfo{author}{Tomczak, J.~M.},
  \bibinfo{author}{Eiben, A.~E.} \& \bibinfo{author}{Brul, S.}
\newblock \bibinfo{journal}{\bibinfo{title}{Population-based parameter
  identification for dynamical models of biological networks with an
  application to saccharomyces cerevisiae}}.
\newblock {\emph{\JournalTitle{Processes}}} \textbf{\bibinfo{volume}{9}},
  \bibinfo{pages}{98} (\bibinfo{year}{2021}).

\bibitem{Storn1997}
\bibinfo{author}{Storn, R.~M.}
\newblock \bibinfo{title}{{Differential evolution—A simple and efficient
  heuristic for global optimization over continuous spaces}}.
\newblock In \emph{\bibinfo{booktitle}{Journal of Global Optimization}},
  \bibinfo{pages}{131--141} (\bibinfo{year}{1997}).

\bibitem{Pedersen2010}
\bibinfo{author}{Pedersen, M.}
\newblock \bibinfo{journal}{\bibinfo{title}{{Good Parameters for Differential
  Evolution}}}.
\newblock {\emph{\JournalTitle{Evolution}}} \bibinfo{pages}{1--10}
  (\bibinfo{year}{2010}).

\end{thebibliography}



\section*{Author contributions statement}

J.L.: Design, implementation, and execution of experiments. Data analysis, data visualization and paper writing. 

K.M.: Experimental design, data analysis, contribution to paper writing. 

J.T.: The manuscript reviewing. 

A.E.E.: Design, Data analysis, contribution to paper writing.

\section*{Competing interests}
The authors declare no competing interests.

\section*{Supplementary Information}

\begin{table}[h]
\caption{Main experiment parameters}
\begin{tabular}{{p{0.18\linewidth} | p{0.08\linewidth}| p{0.63\linewidth}}}
\toprule
Parameters       & Value & Description                                    \\ \midrule
Population size  & ~50    & Number of individuals per generation     \\
Offspring size  & ~25    & Number of offspring produced per generation     \\
Mutation         & ~0.8   & Probability of mutation for individuals        \\ 
Crossover         & ~0.8   & Probability of crossover for individual's body        \\ 
Generations      & ~30   & Termination condition for each run             \\ 
Learning trials  & ~280    & Number of the evaluations performed by RevDE on each robot \\ 
$\mu$ 			 & ~10     & RevDE - Population size 	\\
N 			 & ~30     & RevDE - New candidates per iteration	\\
$\lambda$ 			 & ~10     & RevDE - Top-sample size 	\\
$F$ 			 & ~0.5    & RevDE - Scaling factor 	\\
$CR$  & ~0.9    & RevDE - Crossover probability \\ 
Iterations  & ~10    & RevDE - Number of iterations \\ 
Evaluation time  & ~40    & Duration of evaluation in seconds\\ 
Tournament size  & ~2     & Number of individuals used in the parent selection - (k-tournament)		 \\ 
$\lambda \slash \mu$ & ~0.5    & The ratio used in the survivor selection - ($\mu + \lambda)$  \\
Repetitions      &  ~20    & Number of repetitions per experiment \\ 
\bottomrule 
\end{tabular}
\label{tab:parameters}
\end{table}

\end{document}